\pdfoutput=1
\documentclass[10pt,journal,compsoc]{IEEEtran}

\usepackage{times}
\usepackage{soul}
\usepackage{url}
\usepackage[hidelinks]{hyperref}
\usepackage[utf8]{inputenc}
\usepackage{caption}
\usepackage{graphicx}
\usepackage{amsmath}
\usepackage{amsthm}
\usepackage{booktabs}
\usepackage{algorithm}
\usepackage{algorithmic}
\usepackage{pifont}
\usepackage{multirow}
\usepackage{array}
\usepackage{enumitem}
\usepackage{subcaption}
\usepackage{adjustbox}
\usepackage{graphicx}

\newcommand{\cmark}{\ding{51}}%
\newcolumntype{P}[1]{>{\centering\arraybackslash}p{#1}}
\usepackage{amsfonts}
\usepackage{xcolor}

\newcommand{\Space}[1]{\mathbb{#1}}
\newcommand{\Set}[1]{\mathcal{#1}}
\newcommand{\Rv}[1]{{#1}}
\usepackage{forest}
\usepackage{adjustbox}
\usepackage{caption}
\newcommand{\methodname}[1]{\textbf{#1}} 
\newcommand{\authorname}[1]{\textbf{#1}}

\newtheorem{definition}{Definition}
\newtheorem{assumption}{Assumption}
\theoremstyle{remark}

\newcommand{\tabincell}[2]{\begin{tabular}{@{}#1@{}}#2\end{tabular}}
\newcommand{\indep}{\perp \!\!\! \perp}

\usepackage[sort,numbers,merge]{natbib}

\newcommand{\shortcite}[1]{\cite{#1}}

\begin{document}
\title{Out-Of-Distribution Generalization on Graphs:\\ A Survey}
\author{Haoyang Li,
        Xin Wang,~\IEEEmembership{Member,~IEEE,}
		Ziwei Zhang,~\IEEEmembership{Member,~IEEE,}
		Wenwu Zhu,~\IEEEmembership{Fellow,~IEEE} %
\IEEEcompsocitemizethanks{\IEEEcompsocthanksitem Haoyang Li, Xin Wang, Ziwei Zhang, and Wenwu Zhu are with the Department of Computer Science and Technology in Tsinghua University, Beijing, China. Haoyang Li and Xin Wang contribute equally.\protect\\ E-mail: lihy18@mails.tsinghua.edu.cn, xin\_wang@tsinghua.edu.cn, zwzhang@tsinghua.edu.cn,  wwzhu@tsinghua.edu.cn}%
}

\markboth{Preprint}%
{Li \MakeLowercase{\textit{et al.}}: Bare Demo of IEEEtran.cls for Computer Society Journals}

\IEEEtitleabstractindextext{%
\begin{abstract}
Graph machine learning has been extensively studied in both academia and industry. Although booming with a vast number of emerging methods and techniques, most of the literature is built on the in-distribution hypothesis, i.e., testing and training graph data are identically distributed. However, this in-distribution hypothesis can hardly be satisfied in many real-world graph scenarios where the model performance substantially degrades when there exist distribution shifts between testing and training graph data. To solve this critical problem, out-of-distribution (OOD) generalization on graphs, which goes beyond the in-distribution hypothesis, has made great progress and attracted ever-increasing attention from the research community. In this paper, we comprehensively survey OOD generalization on graphs and present a detailed review of recent advances in this area. First, we provide a formal problem definition of OOD generalization on graphs. Second, we categorize existing methods into three classes from conceptually different perspectives, i.e.,  data, model, and learning strategy, based on their positions in the graph machine learning pipeline, followed by detailed discussions for each category. We also review the theories related to OOD generalization on graphs and introduce the commonly used graph datasets for thorough evaluations. Finally, we share our insights on future research directions. This paper is the first systematic and comprehensive review of OOD generalization on graphs, to the best of our knowledge.
\end{abstract}

}

\maketitle

\IEEEdisplaynontitleabstractindextext

\IEEEpeerreviewmaketitle

\IEEEraisesectionheading{
\section{Introduction}\label{section:introduction}}

\IEEEPARstart{G}{raph}
data is ubiquitous in our daily life.
It has been widely used to model the complex relationships and dependencies between entities, ranging from microscopic particle interactions in physical systems and molecular structures in proteins to macroscopic traffic networks and global communication networks.
Machine learning approaches on graphs, especially for graph neural networks (GNNs), have attracted wide attention and been extensively studied in the last decade.
They have shown great successes in both academia and industry, illustrating their excellent capabilities in a wide range of realistic applications, e.g., social networks~\cite{qiu2018deepinf}, recommendation systems~\cite{wu2020graph}, knowledge representation~\cite{wang2017knowledge}, traffic forecasting~\cite{yu2018spatio}, etc.

Despite the notable success of graph machine learning approaches, the existing literature generally relies on the assumption that the testing and training graph data are drawn from the identical distribution, i.e., the in-distribution (I.D.) hypothesis.
However, in the real world, such a hypothesis is difficult to be satisfied due to the uncontrollable underlying data generation mechanism~\cite{bengio2019meta}. 
In practice, there will inevitably be scenarios with distribution shifts between testing and training graphs~\cite{li2021ood}.
These classic graph machine learning approaches lack the ability of \emph{out-of-distribution} (OOD) generalization, which fail dramatically with significant performance drop under distribution shifts.
Therefore, it is of paramount importance to develop approaches capable of out-of-distribution generalization on graphs, especially for high-stake graph applications, e.g., molecule prediction~\cite{hu2020open}, financial analysis~\cite{yang2020financial}, criminal justice~\cite{agarwal2021towards}, autonomous driving~\cite{liang2020learning}, particle physics~\cite{shlomi2020graph}, as well as pandemic prediction~\cite{panagopoulos2020transfer}, medical detection~\cite{horry2020covid} and drug repurposing~\cite{hsieh2021drug} for COVID-19.

\begin{figure}[t]
	\centering
	\includegraphics[width=1.0\linewidth]{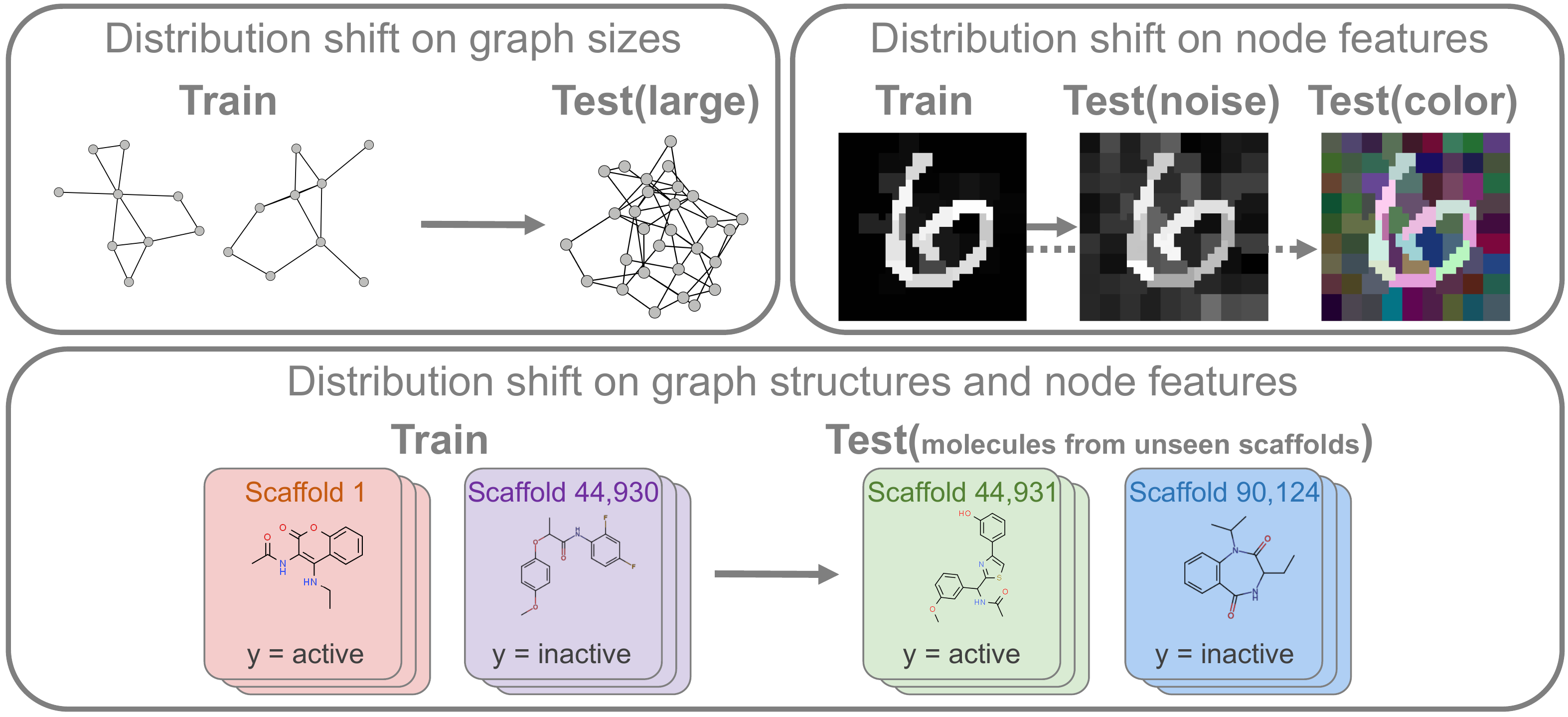}
	\caption{Complex types of distribution shifts on graphs. The distribution shifts can exist on graph sizes, node features, and graph structural properties~\cite{li2021ood}. The OOD generalized graph approaches are expected to perform well on the unseen testing data even under distribution shifts rather than overfitting the training data.}
	\label{fig:distributionshift}
\end{figure}

Out-of-distribution (OOD) generalization algorithm~\cite{shen2021towards, wang2021generalizing, zhou2022domain} aims to achieve satisfactory generalization performance under unknown distribution shifts.
It has been occupying an important position in the research community due to the increasing demand for handling in-the-wild unseen data.
Combining the strength of graph machine learning and OOD generalization, i.e., \textbf{OOD generalization on graphs}, naturally serves as a promising research direction to facilitate graph machine learning model deployments in real-world scenarios.
However, this problem is highly non-trivial due to the following challenges.
\begin{itemize}
	\item \textbf{Uniqueness of graph data}: The non-Euclidean nature of graph-structured data space leads the unique graph model designs and makes obstacles for the direct adoption of OOD generalization algorithms that are mainly developed on Euclidean data (e.g., images and texts).
	\item \textbf{Diversity of graph task}: The problems on graphs are highly diverse, ranging from node-level, link-level to graph-level tasks, along with distinct settings, objectives, and constraints. It is necessary to integrate different levels of graph characterizations into the graph OOD generalization methods.
	\item \textbf{Complexity of graph distribution shift type}: The distribution shifts on graphs can exist on feature-level (e.g., node features) and topology-level (e.g., graph size or other structural properties). Such complex types of graph distribution shifts (as shown in Fig.~\ref{fig:distributionshift}) render more difficulties for OOD generalization.
\end{itemize}
With both opportunities and challenges, it is the right time to review and carry out the studies of graph OOD generalization methods. 
In this paper, we provide a systematic and comprehensive review\footnote{The summary of graph OOD generalization methods reviewed in this survey can be found at \url{https://graph.ood-generalization.com}.} for OOD generalization on graphs for the first time, to the best of our knowledge.
Specifically, to cover the whole life cycle of OOD generalization on graphs, we start by providing a formal problem definition.
We divide the existing methodologies into three conceptually different categories based on their positions in the graph machine learning pipeline, and elaborate typical approaches for each category.
We also review the theories and datasets for evaluations to further promote the research on OOD generalization on graphs.
Last but not least, we share our insights on potential research topics deserving future investigations.

Some related surveys review from the perspectives of graph data augmentation~\cite{ding2022data, zhao2022graph}, graph self-supervised learning~\cite{liu2021graph, xie2022self}, graph adversarial learning~\cite{sun2018adversarial, chen2020survey}, etc. 
However, they are significantly different from ours.
First, they do not focus on the graph OOD generalization that is the center topic of this survey.
Then, a portion of their reviewed methods serves as an important piece of the puzzle for the whole problem of graph OOD generalization. 
To the best of our knowledge, there is no comprehensive review for current advancements of graph OOD generalization methods.

The rest of the paper is organized as follows. In Section~\ref{sec-problem}, we formulate the problem of OOD generalization on graphs and present our categorization of existing literature.
We comprehensively review three categories of methods in Sections~\ref{sec-data}--\ref{sec-optimization}, followed by our review of related theory (in Section~\ref{sec-theory}) and evaluation datasets (in Section~\ref{sec-datasets}). Lastly, we point out future research opportunities in Section~\ref{sec-discussions}.

\begin{figure*}[thbp]
	\centering
	\resizebox{0.9\textwidth}{!}{
	\definecolor{myblue}{RGB}{78, 101, 155}
\definecolor{mybluepurple}{RGB}{138, 140, 191}
\definecolor{mypurple}{RGB}{184, 168, 207}
\definecolor{myyellow}{RGB}{253, 207, 158}
\definecolor{mybrown}{RGB}{182, 118, 108}
\definecolor{amethyst}{rgb}{0.6, 0.4, 0.8}
\definecolor{bananayellow}{rgb}{1.0, 0.88, 0.21}
\definecolor{mypink}{RGB}{231, 188, 198}

\begin{forest}
  for tree={
  grow=east,
  reversed=true,
  anchor=base west,
  parent anchor=east,
  child anchor=west,
  base=left,
  font=\small,
  rectangle,
  draw,
  rounded corners,align=left,
  minimum width=2.5em,
  inner xsep=4pt,
  inner ysep=1pt,
  },
  where level=1{text width=5em,top color=bananayellow!20, bottom color=bananayellow!40}{},
  where level=2{text width=5em,font=\footnotesize,top color=mypink!40, bottom color=mypink!60}{},
  where level=3{font=\footnotesize,yshift=0.26pt,top color=myyellow!50, bottom color=myyellow!70}{},
  [Graph OOD \\ generalization\\methods,top color=amethyst!20, bottom color=amethyst!40
        [Data (Sec.~\ref{sec-data}),text width=6.2em
            [Structure-wise Graph \\ Data Augmentation,text width=8.3em [GAug~\cite{zhao2020data}; MH-Aug~\cite{park2021metropolis};\\ KDGA~\cite{wuknowledge}.
                ]
            ]
            [Feature-wise Graph \\ Data Augmentation,text width=8.3em [GRAND~\cite{feng2020graph}; FLAG~\cite{kong2020flag};\\ LA-GNN~\cite{liu2022local}.
            ]
            ]
            [Mixed-type Graph \\ Data Augmentation,text width=8.3em [GraphCL~\cite{you2020graph}; GREA~\cite{liu2022graph}; DPS~\cite{yu2022finding};\\ AdvCA~\cite{sui2022adversarial}; Mixup~\cite{zhang2017mixup}.
            ]
            ]
        ]
        [Model (Sec.~\ref{sec-model}),text width=6.2em
              [Disentanglement-based\\Graph Models,text width=8.3em [DisenGCN~\cite{ma2019disentangled}; IPGDN~\cite{liu2020independence}; FactorGCN~\cite{yang2020factorizable}; DisC~\cite{fan2022debiasing};
              \\ NED-VAE~\cite{guo2020interpretable}; DGCL~\cite{li2021disentangled}; IDGCL~\cite{li2022disentangled}.
                ]
            ]
            [Causality-based\\Graph Models,text width=8.3em [OOD-GNN~\cite{li2021ood}; StableGNN~\cite{fan2021generalizing}; DGNN~\cite{fan2022debiased}; CAL~\cite{sui2021deconfounded}; DSE~\cite{wu2022deconfounding};\\ CIGA~\cite{chen2022learning}; E-invariant GR~\cite{bevilacqua2021size}; gMPNN$^{\bullet\bullet}$~\cite{zhou2022ood} ; CFLP~\cite{zhao2021counterfactual}; Gem~\cite{lin2021generative}.
              ]
            ]
        ]
        [Learning\\ Strategy (Sec.~\ref{sec-optimization}),text width=6.2em
            [Graph Invariant\\Learning,text width=8.3em [GIL~\cite{li2022gil}; DIR~\cite{shirley2022dir}; GSAT~\cite{miao2022interpretable};  EERM~\cite{wu2022towards}; DIDA~\cite{zhang2022dynamic};\\ SR-GNN~\cite{zhu2021shift}; SizeShiftReg~\cite{buffelli2022sizeshiftreg}; StableGL~\cite{zhang2021stable}.
              ]
            ]
            [Graph Adversarial\\Training,text width=8.3em [DAGNN~\cite{wu2019domain}; GNN-DRO~\cite{sadeghi2021distributionally}; GraphAT~\cite{feng2019graph};\\  CAP~\cite{xue2021cap}; WT-AWP~\cite{wu2022adversarial}; OAD~\cite{wang2021online}.
              ]
            ]
            [Graph Self-supervised\\Learning,text width=8.3em [Pretraining-GNN~\cite{hu2019strategies}; PATTERN~\cite{yehudai2021local}; DR-GST~\cite{liu2022confidence};\\  GraphCL~\cite{you2020graph}; RGCL~\cite{li2022let}; GAPGC~\cite{chen2022graphtta}; GT3~\cite{wang2022test}.
              ]
            ]           
        ]
    ]
\end{forest}
	}
	\caption{Taxonomy of graph OOD generalization methods. We categorize  existing methodologies into three conceptually different branches based on their positions in the graph machine learning pipeline, i.e., data, model and learning strategy. 
	}
	\label{fig-main}
\vspace{-1.5mm}
\end{figure*}
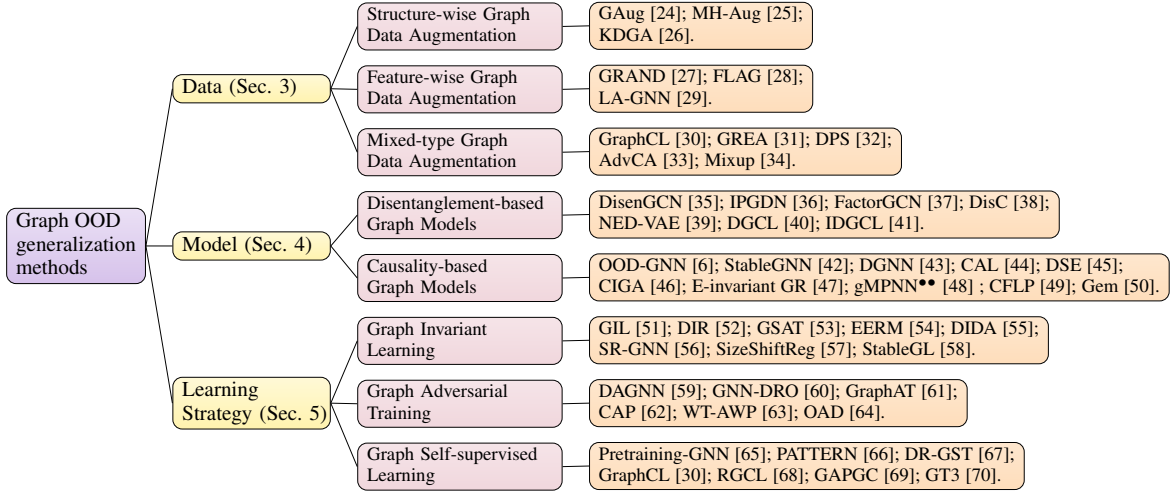

\section{Problem Definition and Categorization}\label{sec-problem}

In this section, we first describe the formulation of OOD generalization on graphs. 
Then we provide the categorization of existing graph OOD generalization methods.

\subsection{Problem Definition}

Let $\Space{X}$ be the input space and $\Space{Y}$ be the label space.
A graph predictor $f_\theta: \Space{X} \rightarrow \Space{Y}$ with parameter $\theta$ maps the input instance $\Rv{X} \in \Space{X} $, i.e., node/link/graph for node-level/link-level/graph-level task, into the label $\Rv{Y} \in \Space{Y}$.
A loss function $\ell$ measures the distance between prediction and ground-truth label.
The graph OOD generalization problem is defined as:
\begin{definition}[Graph OOD generalization] 
Given the training set of $N$ instances (i.e., nodes, links, or graphs) $\Set{D} = \left\{(X_i,Y_i) \right\}_{i=1}^N$ that are drawn from training distribution $P_{train}(\Rv{X}, \Rv{Y})$, where $X_i \in \Space{X}$ and $Y_i \in \Space{Y}$, the goal is to learn an optimal graph predictor $f_\theta^{*}$  that can achieve the best generalization on the data drawn from test distribution $P_{test}(\Rv{X}, \Rv{Y})$, where $P_{test}(\Rv{X}, \Rv{Y}) \neq P_{train}(\Rv{X}, \Rv{Y})$:
\begin{equation}
\label{equ:problemdef}
\small
	f^{*}_\theta = \arg\min_{f_\theta} \Space{E}_{\Rv{X}, \Rv{Y} \sim P_{test}} [\ell(f_\theta(\Rv{X}), \Rv{Y})].
\end{equation}
\end{definition}
The distribution shifts between $P_{test}(\Rv{X}, \Rv{Y})$ and $P_{train}(\Rv{X}, \Rv{Y})$ can lead to the failure of graph predictor built on the in-distribution (I.D.) hypothesis, since directly minimizing the average loss on training instances $\Space{E}_{\Rv{X}, \Rv{Y} \sim P_{train} } [\ell(f_\theta(\Rv{X}), \Rv{Y})]$ can not obtain an optimal predictor that generalizes to testing instances under distribution shifts.
Note that the testing distribution is unknown during the training stage.

\subsection{Categorization}

To tackle the challenges brought by unknown distribution shifts and solve the graph OOD generalization problem, considerable efforts have been made in literature, which can be categorized into three classes:
\begin{itemize}
	\item \textbf{Data}: This category of methods aims to manipulate the input graph data, i.e., graph augmentation. By systematically generating more training samples to increase the quantity and diversity of the training set, graph augmentation techniques are effective in improving the OOD generalization performance. 
	\item \textbf{Model}: This category of methods aims to propose new graph models for learning OOD generalized graph representations, including two types of representative methods: disentanglement-based graph models and causality-based graph models. 
	\item \textbf{Learning Strategy}: This category of methods focuses on exploiting the training schemes with tailored optimization objectives and constraints to enhance the OOD generalization capability, including graph invariant learning, graph adversarial training, and graph self-supervised learning.
\end{itemize}

These three categories of methods solve the graph OOD generalization problem from three conceptually different perspectives. 
We provide the taxonomy in Figure~\ref{fig-main} and elaborate these methods for each category in the following sections.
We also summarize the characteristics of these methods in Table~\ref{table:properties}.

\section{Data}\label{sec-data}

The OOD generalization ability of machine learning models, including graph models, heavily relies on the diversity and quality of training data~\cite{wang2021generalizing}.
In general, the more diverse and high-quality the training data, the better the generalization performance of graph models.
With proper graph augmentation technique, this type of methods can obtain more graph instances with a simple way for training, whose goal can be formulated as:
\begin{equation}
\small
\min_{f_\theta} \Space{E}_{\Rv{X}', \Rv{Y}'} [\ell(f_\theta(\Rv{X}'), \Rv{Y}')],
\end{equation}
where $(\Rv{X}', \Rv{Y}')$ belongs to training set $\Set{D}'$ augmented from $\Set{D}$.
In general, the graph augmentation literature can be summarized into three types of strategies, including \emph{structure-wise augmentations}, \emph{feature-wise augmentations}, and \emph{mixed-type augmentations}.

\subsection{Structure-wise Graph Data Augmentation}
Since the graph structure (i.e., topology) plays an important role in predicting the properties of graphs, some works focus on structure-wise augmentations for the input graphs to generate more diverse training topologies that potentially cover some unobserved testing topologies, leading to better OOD generalization ability.
Here we mainly review the representative graph data augmentation approaches that \emph{claim to or have practically been verified to} improve the OOD generalization in the paper, the same below.
Please refer to the graph augmentation surveys~\cite{ding2022data, zhao2022graph} for more details of other methods.

\methodname{GAug} (Graph Augmentation)~\cite{zhao2020data} proposes to generate augmented graphs via a differentiable edge predictor for improving the generalization. 
It finds that the edge predictors can effectively encode class-homophilic structure to promote intra-class edges and demote inter-class edges in the given graph structure.
Such edge manipulation can not only benefit the prediction accuracy but the generalization ability of the graph models.
GAUG uses an edge prediction module to modify the given input graph for the downstream training and inference processes. It can also learn to generate possible new edges for the input graph. The performance of node-level classification tasks can be improved without any modification at inference time. 
Based on both denoised structure and mimic variability, this graph augmentation achieves a boost in generalization capability.

\methodname{MH-Aug} (Metropolis-Hastings Data Augmentation)~\cite{park2021metropolis} further proposes graph augmentation from a perspective of a Markov chain Monte Carlo sampling~\cite{hastings1970monte} to flexibly control the strength and diversity of augmentation. 
A sequence of augmented samples are drawn from the explicitly designed target distribution that controls the augmentation. 
For tackling the infeasibility of direct sampling from the complex distribution, it adopts the Metropolis-Hastings algorithm to obtain the augmented samples.
Instead of random graph augmentations, this method is more controllable, including an efficient strategy to measure and control the augmentation strength reflecting the structural changes of ego-graphs (or samples in node classification).
Finally, the OOD generalization power of this method is increased by the diverse augmented training samples.

\methodname{KDGA} (Knowledge Distillation for Graph Augmentation)~\cite{wuknowledge} identifies the negative augmentation problem of the graph augmentation methods above, namely these methods could cause overly severe distribution shifts between the augmented graphs for training and the graph for testing, leading to suboptimal generalization. 
KDGA is a graph structure augmentation method proposed based on the knowledge distillation technique to reduce the potential negative effects of distribution shifts. 
Specifically, it extracts the knowledge from the GNN teacher model
trained on the augmented graph data and leverages such knowledge in a partially parameter-shared student model that is tested on the given input graph. The experiments on both homophily and heterophily graph datasets show the effectiveness in node-level tasks.

\subsection{Feature-wise Graph Data Augmentation}
Besides structure-wise augmentations introduced above that remove or add edges for the input graph, some techniques on manipulating node features are also developed recently, showing effectiveness in enhancing the OOD generalization.

\methodname{GRAND} (Graph Random Neural Network)~\cite{feng2020graph} is one simple yet effective feature-wise augmentation method for improving the generalization. It first randomly drops on node features either partially or entirely and then propagates the perturbed node features over the input graph. 
Therefore, each node of the input graph can get rid of the excessive sensitivity to specific neighborhoods that could induce poor OOD generalization. 
Under the homophily assumption on graph~\cite{mcpherson2001birds}, it stochastically creates different
augmented representations for each node. The consistency loss further minimizes the distances of the representations learned from the augmented graphs. 

\methodname{FLAG} (Free Large-scale Adversarial Augmentation on Graphs)~\cite{kong2020flag} is another simple, scalable, and general graph data augmentation method for better GNN generalization. 
It proposes to iteratively augment node features in the input node feature space with gradient-based adversarial perturbations during training, while keeping the graph structures unchanged. It leverages the free adversarial training method~\cite{shafahi2019adversarial} to craft adversarial data augmentations.
Due to its simple and scalable design, this method can conduct efficient training on some large-scale datasets and also can be easily incorporated into the training pipeline of common GNN backbones. 
Different from GRAND that is only designed for tasks on nodes, FLAG can be utilized into node/link/graph level tasks.  

\methodname{LA-GNN} (Local Augmentation for GNN)~\cite{liu2022local} proposes a local augmentation for GNNs to learn the distribution of the node features of the neighbors conditioned on the center node’s feature.
Specifically, it first exploits a generative model to conduct the pre-training for learning the conditional distribution of the neighbors’ node features of the center node’s feature. 
Then, the learned distribution can be used to generate feature vectors associated with the center node as additional input for each training iteration.
Since the pre-training of the generative model and downstream GNN training are decoupled, this data augmentation method is also model-agnostic, which can be applied to most GNN backbones in a
plug-and-play manner.
The feature vectors of new nodes can be directly generated via the generative model, so that it can enhance the generalization of the unseen testing nodes.
The main difference between LA-GNN with some feature-wise graph augmentations above is that it pays more attention to the local information of the node neighbors rather than only focusing on global augmentation concerning the properties of the whole distribution of the graph.

\subsection{Mixed-type Graph Data Augmentation}
Moreover, for combining the advantages of structure-wise and feature-wise graph augmentation methods, some works do not conduct single type of augmentation on graph topology or node feature, but in the mixed-type paradigm, which are increasingly popular in the community for improving OOD generalization.

\methodname{GraphCL} (Graph Contrastive Learning)~\cite{you2020graph} first proposes four general data augmentations for graph-structured data, including node dropping, edge perturbation, attribute masking, and subgraph sampling.
Specifically, node dropping is to randomly remove nodes as well as the links to neighbors.  And the edge perturbation is to randomly add or remove a fraction of edges. Attribute masking is to mask off certain node attributes by setting the attributes to Gaussian noises.
Subgraph sampling is to sample a subgraph using random walk, which includes a fraction of nodes from the input graph.
After obtaining the augmented samples of the  input, it makes the graph encoder maximize representation consistency under augmentations and has shown good OOD generalization ability in graph classification~\cite{ding2021closer}.

\methodname{GREA} (Graph Rationalization Enhanced by Environment-based Augmentations)~\cite{liu2022graph} proposes a data augmentation strategy based on environment replacement to improve the rationale  identification accuracy of the input graphs for OOD generalization. 
The graph rationale is defined as a part of each input graph, i.e., the representative subgraph, that best supports the prediction and can be OOD generalizable.
The authors argue that existing augmentation methods (e.g., GraphCL, etc.) are mainly heuristic modification to the input graphs, which could not directly support the identification of graph rationales. 
They generate an augmented example by replacing the environment subgraph of the input graph with the environment subgraph of another graph and encourage the augmented examples to have the same label of the input graph.
Considering the high complexity of explicit subgraph decoding and encoding, it turns to implicitly conduct rationale-environment separation and representation learning for the original and augmented graphs in latent space.
Based on the accurately identified rationale of the input graph, they verify that the OOD generalization ability is improved.

\methodname{DPS} (Diverse and Predictable Subgraphs)~\cite{yu2022finding} generates several augmented domains (or environments) based on the source domain and further learns consistent semantics among augmented domains and source domain for OOD generalization.
Similar to GREA, it is also a graph data augmentation method which is specific for graph environments to achieve graph OOD generalization.
Since distribution shifts are induced from the disparity between different domains, the graph predictor can generalize to OOD graphs when it performs equally well on multiple domains. 
However, collecting sufficient domains for graph data is usually impractically difficult. So DPS aims to generate augmented domains to tackle the domain scarcity problem which is common on graphs.
Specifically, it adopts several subgraph generators to output diverse subgraphs as augmented domains while maintaining critical information in each subgraph for predicting the graph label. 
For encouraging the diversity of augmented domains, an energy-based regularization is proposed to enlarge the distances between the probability masses of different augmented domains. And for encouraging to learn predictive subgraph, the subgraph generator is equipped with a variational distribution to minimize the risk of the graph predictor.
Based on these diverse domains that preserve consistent predictive semantic information to the source domain, the graph predictor can obtain equal predictive ability across different domains, which can generalize on OOD testing graphs in unseen domains.

\methodname{AdvCA} (Adversarial Causal Augmentation)~\cite{sui2022adversarial} proposes a graph augmentation technique to alleviate the covariate shift problem that is one specific scenario in graph OOD generalization. 
The authors claim that existing graph augmentation strategies suffer from limited environments or unstable causal features, restricting their OOD generalization ability under covariate shift data.
To tackle this problem, AdvCA first proposes two principles for graph augmentation, which are environmental diversity and causal invariance.
The environmental diversity principle encourages the graph augmentation to extrapolate unseen environments (or domains). And the causal invariance principle reduces the distribution gap between the augmented graph data and unseen testing graph data. 
The method consists of two main modules, including adversarial augmenter to adversarially learn the masks on both graph topology and node features for enhancing environmental diversity, causal generator to output the masks that capture causal information. 
Based on the two principles and corresponding designs, AdvCA can get rid of vulnerability under covariate shift.

Besides, in parallel with the development of graph neural networks, \methodname{Mixup} and its variants~\cite{zhang2017mixup, verma2019manifold}, as general data augmentation methods that generate new instances based on the interpolation of the given instances, have been theoretically and empirically shown to improve generalization ability in the fields of computer vision~\cite{zhang2020does} and natural language processing~\cite{guo2020nonlinear}. 
The similar strategies are also applied in graphs~\cite{verma2021graphmix, wang2021mixup, wu2021graphmixup, guo2021intrusion, wang2020nodeaug, han2022g}. 
For example,
\methodname{GraphMix}~\cite{verma2021graphmix} adopts manifold mixup~\cite{verma2019manifold} on node classification tasks by jointly training a fully-connected network (FCN) and a GNN. 
The loss of FCN is computed using manifold mixup while the loss of GNN is computed normally.
A parameter sharing strategy is utilized between the FCN and GNN to help the transfer of critical node representations from the FCN to the GNN.
\methodname{G-Mixup}~\cite{wang2021mixup} interpolates the node features and graph structure in the embedding space as data augmentation, i.e., interpolating the hidden representations of graphs. 
\methodname{NodeAug}~\cite{wang2020nodeaug} analogizes Mixup with a two-branch graph convolution module. It mixes the raw features of a pair of nodes, and feeds them into the two-branch GNN layer, followed by mixing their hidden representations of each layer.
\methodname{ifMixup} (intrusion-free Mixup)~\cite{guo2021intrusion} applies Mixup not for the latent representations but directly on the graph data.
Due to the issue that graph data are irregular and the nodes of two graphs are not aligned, ifMixup assigns indices to the nodes arbitrarily and matches the nodes with the indices.
\methodname{$\mathcal{G}$-Mixup}~\cite{han2022g} tackles the key challenges when mixing up directly on the graph data, as graph data is irregular and not well-aligned, and graph topology between classes is divergent. 
Specifically, it first adopts graphs within
the same class to estimate a graphon. After that, it does not  manipulate graphs directly, but interpolates graphons of different classes in the Euclidean space to obtain the mixed graphons, where the synthetic graphs are produced via sampling based upon the mixed graphons. This method performs well in graph classification datasets with distribution shifts, reflecting its promising OOD generalization.
Since these methods share similar ideas, we use the notation ``Mixup'' to denote these Mixup-based methods that are introduced above in Figure~\ref{fig-main} and Table~\ref{table:properties}.

\begin{table*}[!h]
  \centering
  \small
  \caption{A summary of graph OOD generalization methods. ``Task" denotes the task type that each method focuses on, including node/link/graph level tasks. ``Shift Type" denotes the type of distribution shifts that each method can handle, including topology-level (i.e., graph size and graph structure) and feature-level (i.e., node features) distribution shifts. ``Backbone agnostic" indicates whether the method can be used for other GNN backbones. ``$|\mathcal{E}|>1$" indicates whether the method relies on multiple environments during the training process.
  }
  \label{table:properties}
  \begin{adjustbox}{max width=\textwidth}
  \begin{tabular}{c|c|l|ccc|ccc|cc}
  \toprule
  \multirow{2}{*}{\textbf{Category}}     &      \multirow{2}{*}{\textbf{Subcategory}}    &    \multirow{2}{*}{\textbf{Method}}     & \multicolumn{3}{c|}{\textbf{Task}} & \multicolumn{3}{c|}{\textbf{Shift Type}}    & \textbf{Backbone} & \multirow{2}{*}{$|\mathcal{E}|>1$} \\
                                              &                                                       &                                             & \textbf{Node}   & \textbf{Link}   & \textbf{Graph}  & \textbf{Size}   & \textbf{Structure} & \textbf{Feature} & \textbf{Agnostic}               &                    \\
  \midrule
  \multirow{11}{*}{\tabincell{c}{Data}}                 & \multirow{3}{*}{{\tabincell{c}{Structure-wise \\ Graph Data \\ Augmentation}}} & GAug~\cite{zhao2020data}                     & \cmark &        &        &        & \cmark &        & \cmark &        \\
                                                        &                                                                                & MH-Aug~\cite{park2021metropolis}             & \cmark &        &        &        & \cmark &        & \cmark &        \\
                                                        &                                                                                & KDGA~\cite{wuknowledge}                      & \cmark &        &        &        & \cmark &        & \cmark &        \\ \cmidrule{2-11}
                                                        & \multirow{3}{*}{{\tabincell{c}{Feature-wise \\ Graph Data \\ Augmentation}}}   & GRAND~\cite{feng2020graph}                   & \cmark &        &        &        &        & \cmark & \cmark &        \\
                                                        &                                                                                & FLAG~\cite{kong2020flag}                     & \cmark & \cmark & \cmark &        &      & \cmark & \cmark &        \\
                                                        &                                                                                & LA-GNN~\cite{liu2022local}                   & \cmark &        &        &        &        & \cmark & \cmark &        \\ \cmidrule{2-11}
                                                        & \multirow{5}{*}{{\tabincell{c}{Mixed-type \\ Graph Data \\ Augmentation}}}     & GraphCL~\cite{you2020graph}                  & \cmark &        & \cmark &        & \cmark & \cmark & \cmark &        \\
                                                        &                                                                                & GREA~\cite{liu2022graph}                     &        &        & \cmark &        & \cmark & \cmark & \cmark &        \\
                                                        &                                                                                & DPS~\cite{yu2022finding}                     & \cmark &        & \cmark &        & \cmark & \cmark & \cmark &        \\
                                                        &                                                                                & AdvCA~\cite{sui2022adversarial}              &        &        & \cmark &        & \cmark & \cmark & \cmark &        \\
                                                        &                                                                                & Mixup~\cite{zhang2017mixup}                  & \cmark &        & \cmark &        & \cmark & \cmark & \cmark &        \\
  
  \midrule                                                       
  \multirow{17}{*}{\tabincell{c}{Model}}                & \multirow{7}{*}{{\tabincell{c}{Disentanglement-\\based \\ Graph Models}}}      & DisenGCN~\cite{ma2019disentangled}           & \cmark &        &        &        & \cmark & \cmark &        &        \\
                                                        &                                                                                & IPGDN~\cite{liu2020independence}             & \cmark &        &        &        & \cmark & \cmark &        &        \\
                                                        &                                                                                & FactorGCN~\cite{yang2020factorizable}        &        &        & \cmark &        & \cmark & \cmark &        &        \\
                                                        &                                                                                & DisC~\cite{fan2022debiasing}                 &        &        & \cmark &        & \cmark & \cmark & \cmark &        \\
                                                        &                                                                                & NED-VAE~\cite{guo2020interpretable}          &        &        & \cmark &        & \cmark & \cmark &        &        \\
                                                        &                                                                                & DGCL~\cite{li2021disentangled}               &        &        & \cmark &        & \cmark & \cmark & \cmark &        \\
                                                        &                                                                                & IDGCL~\cite{li2022disentangled}              &        &        & \cmark &        & \cmark & \cmark & \cmark &        \\ \cmidrule{2-11}
                                                        & \multirow{10}{*}{{\tabincell{c}{Causality-\\based \\ Graph Models}}}           & OOD-GNN~\cite{li2021ood}                     &        &        & \cmark & \cmark & \cmark & \cmark & \cmark &        \\
                                                        &                                                                                & StableGNN~\cite{fan2021generalizing}         &        &        & \cmark &        & \cmark & \cmark & \cmark &        \\
                                                        &                                                                                & DGNN~\cite{fan2022debiased}                  & \cmark &        &        &        & \cmark & \cmark & \cmark &        \\
                                                        &                                                                                & CAL~\cite{sui2021deconfounded}               &        &        & \cmark &        & \cmark & \cmark & \cmark &        \\
                                                        &                                                                                & DSE~\cite{wu2022deconfounding}               &        &        & \cmark &        & \cmark &        & \cmark &        \\
                                                        &                                                                                & CIGA~\cite{chen2022learning}                 &        &        & \cmark & \cmark & \cmark & \cmark & \cmark &        \\
                                                        &                                                                                & E-invariant GR~\cite{bevilacqua2021size}     &        &        & \cmark & \cmark &        & \cmark & \cmark &        \\
                                                        &                                                                                & gMPNN$^{\bullet\bullet}$~\cite{zhou2022ood}  &        & \cmark &        & \cmark &        &        & \cmark &        \\
                                                        &                                                                                & CFLP~\cite{zhao2021counterfactual}           &        & \cmark &        &        & \cmark  &        & \cmark &        \\
                                                        &                                                                                & Gem~\cite{lin2021generative}                 & \cmark &        & \cmark &        & \cmark &        & \cmark &        \\
  
\midrule                                                       
\multirow{21}{*}{\tabincell{c}{Learning \\ Strategy}} & \multirow{8}{*}{\tabincell{c}{Graph \\ Invariant \\ Learning}}                 & GIL~\cite{li2022gil}                         &        &        & \cmark &        & \cmark & \cmark & \cmark & \cmark \\
                                                        &                                                                                & DIR~\cite{shirley2022dir}                    &        &        & \cmark &        & \cmark & \cmark & \cmark & \cmark \\
                                                        &                                                                                & GSAT~\cite{miao2022interpretable}            &        &        & \cmark &        & \cmark & \cmark & \cmark &        \\
                                                        &                                                                                & EERM~\cite{wu2022towards}                    & \cmark &        &        &        & \cmark & \cmark & \cmark & \cmark \\
                                                        &                                                                                & DIDA~\cite{zhang2022dynamic}                 &        & \cmark &        &        & \cmark & \cmark & \cmark & \cmark \\
                                                        &                                                                                & SR-GNN~\cite{zhu2021shift}                   & \cmark &        &        &        & \cmark & \cmark & \cmark & \cmark \\
                                                        &                                                                                & SizeShiftReg~\cite{buffelli2022sizeshiftreg} &        &        & \cmark & \cmark &        &        & \cmark &        \\
                                                        &                                                                                & StableGL~\cite{zhang2021stable}              & \cmark &        &        &        & \cmark & \cmark &        & \cmark \\ \cmidrule{2-11}
                                                        & \multirow{6}{*}{\tabincell{c}{Graph \\ Adversarial \\ Training}}               & DAGNN~\cite{wu2019domain}                    &        &        & \cmark &        & \cmark & \cmark &        & \cmark \\
                                                        &                                                                                & GNN-DRO~\cite{sadeghi2021distributionally}   & \cmark &        &        &        &  \cmark &  \cmark & \cmark &        \\
                                                        &                                                                                & GraphAT~\cite{feng2019graph}                 & \cmark &        &        &        & \cmark & \cmark & \cmark &        \\
                                                        &                                                                                & CAP~\cite{xue2021cap}                        & \cmark &        &        &        & \cmark & \cmark & \cmark &        \\
                                                        &                                                                                & WT-AWP~\cite{wu2022adversarial}              & \cmark &        & \cmark &        & \cmark & \cmark & \cmark &        \\
                                                        &                                                                                & OAD~\cite{wang2021online}                    & \cmark &        &        &        & \cmark & \cmark & \cmark &        \\ \cmidrule{2-11}
                                                        & \multirow{7}{*}{\tabincell{c}{Graph \\ Self-supervised \\ Learning}}           & Pretraining-GNN~\cite{hu2019strategies}      &        &        & \cmark &        & \cmark & \cmark & \cmark &        \\
                                                        &                                                                                & PATTERN~\cite{yehudai2021local}              &        &        & \cmark & \cmark &        &        & \cmark &        \\
                                                        &                                                                                & DR-GST~\cite{liu2022confidence}              & \cmark &        &        &        & \cmark & \cmark & \cmark &        \\
                                                        &                                                                                & GraphCL~\cite{you2020graph}                  & \cmark &        & \cmark &        & \cmark & \cmark & \cmark &        \\
                                                        &                                                                                & RGCL~\cite{li2022let}                        &        &        & \cmark &        & \cmark & \cmark & \cmark &        \\
                                                        &                                                                                & GAPGC~\cite{chen2022graphtta}                &        &        & \cmark &        & \cmark & \cmark & \cmark &       \\
                                                        &                                                                                & GT3~\cite{wang2022test}                &        &        & \cmark &        & \cmark & \cmark & \cmark &       \\

\bottomrule
\end{tabular}
\end{adjustbox}
\end{table*}

\section{Model}\label{sec-model}
Besides augmenting the input graph data to assist achieving good OOD generalization, there are branches of works that specially design new graph models, i.e., $f_\theta$ in Eq. \eqref{equ:problemdef}.
By introducing some prior knowledge to model design, the graph model is endowed with the ability to produce graph representation with the properties that could help to improve OOD generalization.
Along this branch, there are two kinds of popular techniques: \emph{disentanglement-based graph models} and \emph{causality-based graph models}.

\subsection{Disentanglement-based Graph Models}
In this section, we introduce the graph models based on disentanglement for OOD generalization.

The formation of a real-world graph typically follows a complex and heterogeneous process driven by the interaction of many latent factors.
Disentangled graph representation learning aims to learn representations that separate these distinct and informative factors behind the graph data and characterize these factors in different parts of the factorized vector representations~\cite{ma2019disentangled}. 
Such representations have been demonstrated to be good representations, 
and able to benefit OOD generalization~\cite{wang2022disentangled, montero2020role, dittadi2020transfer}. 
The existing methods fall into three groups, including supervised disentanglement methods~\cite{ma2019disentangled, liu2020independence, yang2020factorizable, fan2022debiasing}, unsupervised generative disentanglement methods~\cite{guo2020interpretable}, and self-supervised contrastive disentanglement methods~\cite{li2021disentangled, li2022disentangled}.

\methodname{DisenGCN}~\cite{ma2019disentangled} is the first method to learn disentangled node representations, whose key ingredient is a disentangled multichannel convolutional layer DisenConv.
Executing inside DisenConv, the proposed neighborhood routing mechanism is to identify the factor that may cause the link from a center node to one of its neighbors, and accordingly send the neighbor to the channel responsible for that factor.
It infers the latent factors by iteratively analyzing the potential subspace clusters formed by the node and its neighbors,
after projecting them into several subspaces.
The authors prove that after a sufficient number of iterations, the proposed neighborhood routing mechanism can converge.
Therefore, each channel of DisenConv can extract features specific to only one disentangled latent factor from the neighbor nodes, and perform a convolution operation independently.
By stacking multiple DisenConv layers, DisenGCN is able to extract information beyond the local neighborhood and produce disentangled representations.
Since the latent factors of nodes are disentangled, it could lead to better OOD generalization performance.

\methodname{IPGDN} (Independence Promoted Graph Disentangled Network)~\cite{liu2020independence} extends DisenGCN~\cite{ma2019disentangled} by explicitly encouraging the latent factors to be as independent as possible in addition to the neighborhood routing mechanism for disentangling latent factors behind graphs. It minimizes the dependence among different representations with a kernel-based measure Hilbert-Schmidt Independence Criterion (HSIC)~\cite{gretton2007kernel}. 
Specifically, to disentangle the target node, the convolution layer of IPGDN first constructs features from different aspects of its neighbors via disentangled representation learning, and then encourages the independence among latent representations through minimizing HSIC to obtain the final results.
Note that the disentangled representation learning and independence regularization are jointly optimized in a unified framework, leading to more disentangled representations when compared with DisenGCN.
And both DisenGCN~\cite{ma2019disentangled} and IPGDN~\cite{liu2020independence} are proposed for handling node-level tasks on graphs.

\methodname{FactorGCN} (Factorizable GCN)~\cite{yang2020factorizable} is a disentangled GNN model for graph-level representation learning.
It adopts a factorizing mechanism by decomposing input graphs into several interpretable factor graphs for graph-level disentangled representations.
Each of the factor graphs is separately sent to a GCN, tailored to aggregate features in terms of only one disentangled latent factor, followed by an aggregating operation that concatenates together all derived features of disentangled latent factors. 
The final produced graph-level representations present block-wise interpretable features, and each of the factorized representations corresponds to a disentangled and interpretable relation space.
These steps constitute one layer of FactorGCN, so that FactorGCN can produce a hierarchical disentanglement with various numbers of factor graphs at different levels by stacking a number of layers to disentangle the
input data at different levels. 

Compared with the methods disentangling latent factors, \methodname{DisC} (Disentangled Causal Substructure)~\cite{fan2022debiasing} is a disentangled GNN model directly disentangling causal and noncausal information of the input graph. By explicitly disentangling the input graph into causal and bias subgraphs, this method can only utilize the causal substructures to make stable predictions when severe bias appears under distribution shifts. 
Specifically, it first filters edges into causal and bias (i.e., noncausal) subgraphs by a parameterized edge mask generator, whose parameters are shared across entire datasets. The edge masker is expected to indicate the importance for each edge and extract causal and bias subgraphs. 
Then, the causal and bias subgraphs are fed to two GNNs trained with causal-aware weighted cross-entropy loss and bias-aware generalized cross-entropy loss respectively, leading to disentangled representations.
Next, it further permutes the latent representations extracted from different graphs to generate more training samples. Although containing both causal and bias information, the causal and bias subgraph of newly generated samples are decorrelated. 
Finally, the proposed model could focus on the true correlation between the disentangled causal subgraphs and labels for achieving OOD generalized prediction.

Besides the supervised methods above, there exist some unsupervised disentangled methods.

\methodname{NED-VAE} (Node-Edge Disentangled Variational
Auto-encoder)~\cite{guo2020interpretable} is a deep unsupervised generative approach for disentanglement learning on graphs, which can automatically capture the independent latent factors in both edges and nodes from attributed graphs. 
The objective is designed for node-edge joint disentanglement by optimizing three sub-encoders (i.e., a node encoder, an edge encoder, and a node-edge co-encoder) that learn the three types of representations, and two sub-decoders (i.e., a node-decoder and an edge decoder) that co-generate both nodes and edges to model the complicated relationships between nodes and edges. 
The base NED-VAE can also be extended to realize the group-wise and variable-wise disentanglement to support more fine-grained disentanglement.

Since reconstruction in unsupervised generative methods could be computationally expensive and even introduce bias that has a negative effect on the learned representations, \methodname{DGCL} (Disentangled Graph Contrastive Learning)~\cite{li2021disentangled} first proposes to learn disentangled graph representations with self-supervision. 
Specifically, it first identifies the latent factors behind the input graph and derives its factorized representations by the tailored disentangled graph encoder whose key ingredient is a multi-channel message-passing layer. Each of the factorized representations describes a latent and disentangled aspect pertinent to a specific latent factor of the graph. 
Then it conducts factor-wise contrastive learning in each representation subspace characterized by each factor independently instead of in the whole representation space. This tailored design can encourage that each disentangled factor of the factorized representations is sufficiently discriminative only under one specific aspect of the whole graph, so as to help the graph encoder produce disentangled graph representations that independently reflect the expressive information of latent factors.
Unlike generative models, contrastive learning is an instance-wise discriminative approach that aims at making similar instances closer and dissimilar instances far from each other in representation space~\cite{jaiswal2020survey, le2020contrastive}, so that it can get rid of computationally expensive graph reconstruction and learn informative graph representations.

To further promote the disentanglement of the learned graph representations, \methodname{IDGCL} (Independence Promoted Disentangled Graph Contrastive Learning)~\cite{li2022disentangled} further extends DGCL by explicitly employing HSIC~\cite{gretton2007kernel} to eliminate the dependence among disentangled representations that reflect different aspects of graphs pertinent to different latent factors.
Since  the disentangled graph representations are expected to capture mutually exclusive information in terms of the latent factors, IDGCL formulates the statistical independence among different latent representations effectively. 
The factor-wise contrastive representation learning and independence regularization are jointly optimized in a unified framework, so that the disentangled graph encoder can produce better disentangled graph representations. Compared with the existing methods, IDGCL encodes a graph with multiple disentangled representations in self-supervised manner, making it possible to explore the meaning of each channel, which benefits in more explainability and OOD generalization for producing graph representations.

\subsection{Causality-based Graph Models}

In this section, we introduce the graph models based on causality for OOD generalization.

Causal inference is one important technique to achieve OOD generalization.
Graph machine learning models tend to exploit subtle statistical correlation existing in the training set even though it is a spurious correlation (unexpected ``shortcut") for predictions to boost training accuracy. 
The performance of graph models that heavily rely on the spurious correlations can be substantially degraded since the spurious correlations could change in the wild OOD testing environments.
In contrast, the causality-based graph models supported by causal inference theory can inherently capture causal relations between input graph data and labels that are stable under distribution shifts~\cite{scholkopf2021toward}, leading to good OOD generalization.
The existing methods can be divided according to their theoretical ground including confounder balancing~\cite{li2021ood, fan2021generalizing, fan2022debiased}, predefined structural causal model~\cite{sui2021deconfounded, wu2022deconfounding, bevilacqua2021size, zhou2022ood}, and counterfactual inference~\cite{zhao2021counterfactual} and Granger causality~\cite{lin2021generative}.

\subsubsection{Confounder Balancing based Methods}

Some methods~\cite{li2021ood, fan2021generalizing, fan2022debiased} introduce confounder balancing into graph models.

\methodname{OOD-GNN}~\cite{li2021ood}, backed by confounder balancing theory~\cite{kuang2018stable} in causality, first tackles the OOD generalization problem by a non-linear decorrelation operation on graphs. 
Specifically, OOD-GNN proposes to eliminate the statistical dependence between causal and noncausal graph representations of the graph encoder by a nonlinear graph representation decorrelation method utilizing random Fourier features~\cite{rahimi2007random}, which scales linearly with the sample size and can get rid of spurious correlations. 
The parameters of the graph encoder and sample weights for graph representation decorrelation are optimized iteratively to learn discriminant graph representations for predictions.
Note that, the decorrelation operation actually has the same effect with confounder balancing that encourages the independence between treatment and confounder.
The graph encoder trained on the weighted dataset can estimate the causal effect of the variables in graph representations to the labels more accurately, while getting rid of the spurious correlations.
In this way, OOD-GNN achieves the satisfactory performance on several graph benchmarks with various types of distribution shifts (i.e., shifts on graph sizes, node features, and graph structures), indicating its strong OOD generalization ability in the wild environments.

\methodname{StableGNN}~\cite{fan2021generalizing} proposes to exploit a differentiable graph pooling layer to extract subgraph-based decorrelated representations based on sample reweighting, which is similar in principle to OOD-GNN.
First, the graph high-level variable learning component employs a graph pooling layer~\cite{xu2018powerful,ying2018hierarchical} to map nearby low-level nodes to a set of clusters, where each cluster is expected to be one densely-connected subgraph unit of original graph. 
Then, it generates the cluster-level embeddings through aggregating the node embeddings in the same cluster, and aligns the cluster semantic space across graphs through an ordered concatenation operation. The cluster-level embeddings act as the high-level variables for graphs.
Next, the sample weights are optimized to eliminate the statistical dependences between these high-level variables.
Thus, the graph encoder can concentrate more on the true connection between discriminative substructures and labels, leading to good OOD generalization ability.

In addition to the graph-level decorrelation models above, \methodname{DGNN} (Debiased GNN)~\cite{fan2022debiased} is a node-level decorrelation model with a similar methodology with StableGNN~\cite{fan2021generalizing} that removes the spurious correlations on nodes to achieve stable predictions under distribution shifts. Specifically, it proposes a framework for OOD generalized node representation learning by jointly optimizing a decorrelation regularizer and a weighted GNN model. The decorrelation regularizer is expected to learn a set of sample weights for eliminating the spurious correlation between causal and noncausal node information for OOD generalization. And the learned sample weights via the decorrelation regularizer are used to reweight the prediction loss of GNN model so that the prediction could be OOD generalized.

\subsubsection{Structural Causal Model based Methods}

Some methods~\cite{sui2021deconfounded, wu2022deconfounding, bevilacqua2021size, zhou2022ood, chen2022learning} take the structural causal model (SCM) into account in their model designs.
In general, the SCM describes the underlying causal mechanisms.
It can improve OOD generalization when introducing appropriate causal mechanisms into model designs.

\methodname{CAL} (Causal Attention Learning)~\cite{sui2021deconfounded} takes a causal look at the GNN model and constructs a structural causal model via presenting the causality among five variables: graph data, causal feature, shortcut feature, graph representation, and prediction. Based on this SCM, they focus on the backdoor path between causal feature $C$ and prediction, wherein the shortcut feature $S$ plays a confounder role. 
This backdoor path could form spurious correlation, namely using the shortcut feature instead of using causal feature to make predictions, leading to poor OOD generalization under distribution shifts.
Therefore, this method exploits the do-calculus on the causal feature to cutting off the backdoor path (i.e., backdoor adjustment~\cite{glymour2016causal}), and gets rid of the confounding effect. 
Finally, it can learn the true relationships between the causal feature and prediction, without being influenced by the unstable shortcut features, which enhances OOD generalization on graph classification tasks.

\methodname{DSE} (Deconfounded Subgraph Evaluation)~\cite{wu2022deconfounding} proposes to faithfully measure the causal effect of explanatory subgraphs on the prediction. The authors claim that distribution shift is hardly measurable, so that it is hard to block the backdoor path from causal subgraph to label by the backdoor adjustment given the predefined SCM. So, they utilize front-door adjustment and introduce a surrogate variable of the causal subgraphs. 
Instead of adopting the feature removal principle that is used in assessing the explanatory subgraph, it designs a generative model, termed conditional variational graph auto-encoder, to generate the possible surrogates that conform to the data distribution. 
Therefore, it can conduct unbiased estimation of the relation between causal subgraph and label. 
Since evaluating the explanatory causal subgraphs unbiasedly, it mitigates the out-of-distribution effect and achieves good OOD generalization.

\methodname{CIGA} (Causality Inspired Invariant Graph Learning)~\cite{chen2022learning} further categorizes the latent interaction between causal part $C$ and noncausal part $S$ into fully informative invariant features (FIIF) and partially informative invariant features (PIIF), depending on whether the latent causal part $C$ is fully informative about label $Y$, i.e.,$(S, E) \indep Y | C$. For FIIF assumption, the noncausal part $S$ is directly controlled by the causal part $C$. And for PIIF, the noncausal part $S$ is indirectly controlled by the causal part $C$ through the label $Y$. The two SCMs exhibit different behaviors in the observed distribution shifts. If one of FIIF or PIIF is excluded, the performances of graph OOD generalization can degrade dramatically. 
Similarly, CIGA instantiates the causal part $C$ as the critical subgraph that includes the information about the underlying causes of the label. So the OOD generalization can be achieved by identifying this critical subgraph that maximally preserves the intra-class information among different training environments, hence the predictions will be stable to distribution shifts.

\methodname{E-invariant GR}~\cite{bevilacqua2021size}
proposes a twin network directed acyclic
graph~\cite{balke2011probabilistic} as their SCM to learn size-invariant graph representations (GR) that better extrapolate between test and train graph data. Different from the SCMs mentioned above, the proposed SCM depicts the more complex and fine-grained relations among several variables, including graphon, train/test environment, node feature, edge, and graph size. 
In this SCM, the training graph is characterized by a graphon, which defines both the label and structural and attribute characteristics of graphs. The training environment is indicated by one unobserved environment variable that represents specific graph properties in terms of environments, so that it could change between the training and test set. 
Based on this SCM, the authors propose approximately size-invariant graph representation that is able to extrapolate to OOD test data and prove that the learned graph representation can perform no worse on the OOD test data than on a test dataset having the same environment distribution as the training data.
Furthermore, this method can achieve extrapolations based on only one training environment (e.g., all training graphs have the same size).

Since E-invariant GR~\cite{bevilacqua2021size} only studies the OOD generalization of GNNs for graph classification, 
\methodname{gMPNN$^{\bullet\bullet}$}~\cite{zhou2022ood} 
further extends it to study the OOD generalization of GNNs for link prediction in a similar setting, where test graph sizes are larger than training graphs. 
Specifically, the authors first proposed a SCM assuming the data generation process for the goal to learn link predictors that generalize under distribution shifts on graph sizes.
And they prove nonasymptotic bounds to indicate that as the sizes of test graphs increase, the link predictors based on permutation-equivariant structural node embeddings will converge to a random guess. 
They show that the output structural pairwise embeddings can converge to embeddings of a continuous function that achieves OOD generalization in link prediction tasks.

\subsubsection{Counterfactual Inference and Granger Causality based Methods}

Besides, some graph OOD methods are inspired by counterfactual learning~\cite{glymour2016causal}, which is at the highest level in the causation ladder~\cite{pearl2018book} and answers what would happen in another possible world if something had or had not happened.
And some methods are motivated by Granger causality~\cite{granger1969investigating}, which describes a causal relationship between variables of some feature and label if we are better able to predict label using all available information than if the information apart from such feature had been used.

\methodname{CFLP} (Counter-Factual Link Prediction)~\cite{zhao2021counterfactual} focuses on OOD link prediction tasks to learn the causal relationship between the global graph structure and link existence by training GNN-based link predictors to predict both factual and counterfactual links. It aims to deal with the counterfactual question: “would the link still exist if the graph structure became different from observation?” By answering this question, the counterfactual links will be used to train the graph encoder for producing OOD generalized representation. 
To generate counterfactual link samples, this method employs causal models that treat the information (i.e., learned representations) of node pairs as context, global graph structural properties as treatment, and link existence as outcome. After that, the proposed model can generate counterfactual training link samples and thus learn representations from both the factual (i.e., observed) and counterfactual (i.e., generated) links for improving OOD generalization.

\methodname{Gem}~\cite{lin2021generative}, built upon the Granger causality, inputs the original computation graph into the explainer and outputs a causal explanation graph, exhibiting better generalization abilities. 
This method considers there exists a causal relationship between this edge/node and its corresponding prediction if the prediction performance decreases as some node or edge is missing.  
Since graph data is inherently interdependent, where nodes and their edges are correlated variables, it further incorporates various graph rules, e.g., connectivity check, to encourage the obtained explanations to be valid and human-intelligible causal subgraphs.
Finally, this method can provide interpretable causal explanations and OOD generalized predictions for GNNs.

\section{Learning Strategy}\label{sec-optimization}

Besides graph data augmentation and graph models, some works focus on exploiting training schemes with tailored optimization objectives and constraints to promote OOD generalization, including graph invariant learning, graph adversarial training, and graph self-supervised learning.

\subsection{Graph Invariant Learning}

First, we introduce the graph invariant learning methods for OOD generalization.

Invariant learning, which aims to exploit the invariant relationships between features and labels across different distributions while disregarding the variant spurious correlations, can provably achieve satisfactory OOD generalization under distribution shifts~\cite{arjovsky2019invariant,chang2020invariant,ahuja2021invariance}.
When assessing causality is challenging or the strong assumptions are potentially violated in practice, it can approximate the task by searching features that are invariant under distribution shifts~\cite{chang2020invariant} for OOD generalization.
Invariant learning assumes that the information of each instance for prediction includes two parts, i.e., invariant part whose relationship with the label is stable across different environments, and variant part whose relationship with the label can change across different environments. 
A good OOD generalization can be obtained when making predictions only on the invariant information. 
Along this line, there are mainly two types of graph invariant learning methods: invariance optimization~\cite{li2022gil, shirley2022dir, wu2022towards, zhang2022dynamic, miao2022interpretable} and explicit representation alignment~\cite{zhu2021shift, buffelli2022sizeshiftreg, zhang2021stable}.

\subsubsection{Invariance Optimization}
These methods are built upon the invariance principle to address the graph OOD generalization problem.
The invariance principle assumes the invariance property inside the data, so that it can find such invariance in multiple environments to achieve OOD generalization. The assumption can be formulated as:
\begin{assumption} (Invariance Assumption). There exists a portion of information $\Phi(X)$ inside input instance $X$ such that $\forall e, e^\prime \in \mathrm{supp}(\mathcal{E}), P^e(Y|\Phi(X)) = P^{e^\prime}(Y|\Phi(X))$, where $\mathcal{E}$ denotes all possible environments and $\Phi(X)$ is often called as invariant rationales of input instance $X$.
\end{assumption}

Following the recent invariant learning based OOD generalization studies~\cite{arjovsky2019invariant, chang2020invariant, ahuja2021invariance}, these invariance optimization methods treat the cause of distribution shifts between testing and training graph data as a potential unknown environmental variable $\Rv{e}$. The optimization objective can be formulated as:
\begin{equation}
\small
\min_{f_\theta} \max_{e\in \mathrm{supp}(\mathcal{E})} \mathcal{R}(f_\theta|e),
\end{equation}
where $\mathcal{R}(f_\theta|e)=\mathbb{E}_{\Rv{X},\Rv{Y} \sim P^e}[\ell(f_\theta(\Phi(\Rv{X})),\Rv{Y})]$ is the risk of the $f_\theta$ on the environment $e$ that makes predictions based on the invariant information $\Phi(X)$.
Therefore, as shown in the last column of Table~\ref{table:properties}, this type of methods relies on explicit multiple-environment split (indicated by $|\mathcal{E}| > 1$) that can be provided in advance or generated during the training process.

\methodname{GIL} (Graph Invariant Learning)~\cite{li2022gil} is proposed to capture the invariant relationships between predictive graph structural information (i.e., subgraphs or rationales) and labels under distribution shifts for graph-level OOD generalization.
One of the main challenges for graph invariant learning is that the environment labels for graphs is generally unobserved or prohibitively expensive to collect, leading that it is difficult to learn invariance in multiple environments. Therefore, this method first studies invariant learning without explicit environment split.
Specifically, GIL jointly optimizes three mutually promoting modules, including the invariant subgraph identification module, the environment inference module, and the invariant learning module. 
First, the invariant subgraph identification module is a GNN-based subgraph generator $\Phi(\cdot)$. Given the input graph $G$, it identifies the invariant subgraph $\Phi(G)$ and defines the rest of the graph, i.e., the complement of invariant subgraph, as the variant subgraph and denote it as $G\backslash \Phi(G)$.
Then, the environment inference module cluster all identified variant subgraphs of the datasets to infer the latent environments. The intuition is that since the invariant subgraph captures invariant relationships between predictive graph structural information and labels, the variant subgraphs in turn capture variant correlations under different distributions, which are environment-discriminative features. 
Finally, the invariant learning module optimizes the proposed maximal invariant subgraph generator criterion given the identified invariant subgraphs and inferred environments to generate graph representations capable of OOD generalization under distribution shifts. 
Theories are provided to show that the OOD generalization problem on graphs is equivalent to finding a maximal invariant subgraph generator of GIL, and further prove that GIL satisfies permutation invariance.

\methodname{DIR} (Discovering Invariant Rationale)~\cite{shirley2022dir} is proposed to handle graph-level OOD generalization tasks by discovering invariant subgraphs $\Phi(G)$ for GNN under interventional distributions.
The basic setting of DIR is also different from the traditional setting where environments are observable and attainable, but follows a similar setting with GIL that does not assume explicit environment split in advance.
In detail, it uses a GNN-based subgraph generator to split the input graph into invariant and variant subgraphs under distribution shifts, which are encoded by the encoder into representations respectively. 
Then, the proposed distribution intervener conducts interventions on the variant representations to create multiple interventional distributions as the multiple environments.
Finally, the two classifiers that are respectively built upon the invariant and variant subgraphs make predictions for the input graph instance jointly, so that the invariant risk is minimized across different environments. 
With this strategy, DIR can capture the invariant rationales that are stable across different distributions while filtering out the spurious patterns that are unstable for OOD generalization.

\methodname{GSAT} (Graph Stochastic Attention)~\cite{miao2022interpretable} addresses graph-level OOD generalization problem utilizing the attention mechanism to build inherently interpretable GNNs for learning invariant subgraphs $\Phi(G)$ under distribution shifts. 
The learned invariant subgraphs of GSAT root in the notion of information bottleneck~\cite{tishby2015deep}. 
The attention is formulated as the information bottleneck by injecting stochasticity into the attention mechanism so as to constrain the information flow from the input graph to the prediction.
The injected stochasticity over the invariant label-relevant subgraphs can be automatically reduced during the training stage, while that over the variant label-irrelevant subgraphs can be kept.
Besides, GSAT also penalizes the amount of information from the input graph data.
Finally, GSAT can output the interpretable and OOD generalizable subgraphs that provably do not contain patterns that are spuriously correlated with the task under some assumptions.
Note that GSAT is also compatible with pre-trained models and further improves the performances.

Besides the graph-level OOD generalized methods, \methodname{EERM} (Explore-to-Extrapolate Risk Minimization)~\cite{wu2022towards} is designed to handle node-level tasks under distribution shifts, which can achieve a valid solution for the node-level OOD problem under mild conditions.
First, to  account for the non-IID nature of nodes on graphs, this method  proposes to transform a graph into a set of ego-graphs for center nodes, so that it can formulate the node-level OOD generalization problem inspired by the graph-level problem. 
Then, it extends the invariance principle with the recursive computation on the induced BFS trees of ego-graphs to consider the structural information. 
Finally, the GNN backbone in EERM is optimized by minimizing the mean and variance of risks from multiple training environments that are generated by the environment generators, while the environment generators are trained by maximizing the variance loss via a policy gradient method.
The authors also derive an upper bound of EERM on the OOD
error which can be effectively controlled when optimizing the training objective.

\methodname{DIDA} (Disentangled Intervention-based Dynamic Graph Attention Network)~\cite{zhang2022dynamic} is the first method to handle graph OOD generalization under more complex spatial-temporal distribution shifts. 
The existing methods usually focus on only spatial distribution shifts existing on node features or graph structures while can not be directly utilized in more complex scenarios where the distribution shifts can simultaneously exist in spatial and temporal information. 
Specifically, it first designs a disentangled spatial-temporal attention network to discover the invariant and variant patterns behind the dynamic graphs, which enables each node to attend to all its historic neighbors through a disentangled attention message-passing mechanism.
Then, it introduces a spatial-temporal intervention mechanism to create multiple intervened distributions via sampling and reassembling the variant patterns across neighborhoods and time, leading that the spurious correlations between the variant patterns and labels can be eliminated.
Note that the variant patterns are highly entangled across nodes and it is computationally expensive if directly generating and mixing up subsets of structures and features to do intervention. So, this method approximates the intervention process with summarized patterns obtained by the disentangled spatio-temporal attention network instead of the original structures and features.
Lastly, the invariance regularization is used to minimize prediction variance in multiple intervened distributions.
Therefore, this method can capture and utilize invariant patterns with stable predictive abilities to labels for making predictions under spatial-temporal distribution shifts. 

\subsubsection{Explicit Representation Alignment}

The key idea of this line of works is to explicitly align the graph representations among multiple environments (or domains) to learn environment-invariant graph representations for OOD generalization.
The graph representation alignment strives to minimize the difference (or encourage the similarity) across multiple environments via the introduced regularization strategy, which can be formulated as: 
\begin{equation}
    \small
    \min_{f_\theta} \Space{E}_{\Rv{X}, \Rv{Y}} [\ell(f_\theta(\Rv{X}), \Rv{Y})] + \ell_{reg}(\mathcal{E}),
    \end{equation}
    where $\ell_{reg}(\mathcal{E})$ denotes the loss of the adopted regularizer.
And the multiple environments $\mathcal{E}$ for calculating the regularizer are also usually unavailable in advance for most graph scenarios and are generated during the training process.

\methodname{SR-GNN} (Shift-Robust GNN)~\cite{zhu2021shift} proposes to address node-level OOD generalization in GNNs by explicitly minimizing the distributional differences between biased training data and a graph's true inference distribution of graphs.
It encourages a biased sample of labeled nodes to more closely conform to the distributional characteristics present in an independent and identically distributed sample of the graph.
The two kinds of bias occurring in both deeper GNNs and more recent linearized (shallow) versions of these models can be handled.
Specifically, SR-GNN first addresses the distribution shift via a regularization over the hidden layers of the network for standard GNN models (e.g., GCN~\cite{kipf2016semi}) that iteratively update information upon the graph structure. 
The regularizations for measuring discrepancy among different distributions can be maximum mean discrepancy (MMD)~\cite{long2015learning} or central moment discrepancy (CMD)~\cite{zellinger2019robust}.
Then, for the linearized models (e.g., SimpleGCN~\cite{wu2019simplifying}) that decouple GNNs into non-linear feature encoding and linear message passing, SR-GNN adopts an instance reweighting strategy for encouraging the training examples to be representative over the graph data, since the graph can introduce bias over the features after all learnable layers.
It learns a group of optimal instance weights via kernel mean matching (KMM)~\cite{gretton2009covariate}.

\methodname{SizeShiftReg}~\cite{buffelli2022sizeshiftreg} aims to train GNNs with good size generalization performance from smaller to larger graphs, which adopts a similar idea with SR-GNN~\cite{zhu2021shift}. 
It does not rely on handcrafting GNNs based on specific knowledge or assumptions, but studies a general regularization for any GNNs to be OOD generalizable to the graph size distribution shifts. 
The introduced graph coarsening strategy is to simulate the distribution shifts in the size of the training graphs. And the proposed regularization is expected to encourage the GNNs to be OOD generalized. 
For a given training graph, they minimize the discrepancy measured by CMD~\cite{zellinger2019robust} between the distributions of the node representations learned by the GNNs from the original training graphs and the coarsened graphs. 
Under such a training paradigm, the learned GNNs can achieve good OOD generalization among different coarsened versions of the graph as well as graphs with unknown size.

\methodname{StableGL}~\cite{zhang2021stable} focuses on stable graph learning (GL) to capture environment-invariant node properties and explicitly balance the multiple environments for generalizing well under distribution shifts.
Given one input graph as the training environment, they aim to train a GNN that has a high average prediction performance but a low variance of performance on multiple agnostic testing environments.
In more detail, the proposed method first performs biased selection on the input training graph to construct multiple training environments.
From a local perspective, since one node in graph is partially represented by the other neighbor nodes, this method proposes to capture stable node properties via reweighting the neighborhood aggregation process. 
From a global perspective, the authors find that the prediction errors in different environments progressively diverge in biased training, eventually leading to unstable performance across environments.
Therefore, the proposed method explicitly aligns the training process by reducing the training gap among different training environments, enforcing the learned GNN to generalize well across unseen testing environments. Different from SR-GNN~\cite{zhu2021shift} and SizeShiftReg~\cite{buffelli2022sizeshiftreg} that adopt some discrepancy measurement like MMD or CMD, the regularization in this method is directly to minimize the variance of training losses in several environments.

\subsection{Graph Adversarial Training}

In this section, we discuss the graph adversarial learning methods for OOD generalization.

Adversarial training
has been demonstrated to improve model robustness against adversarial attacks and OOD generalization ability.
Here we mainly focus on the graph adversarial training methods that improve the generalization ability, while the works protecting GNNs from attacks can be found in the previous survey~\cite{sun2018adversarial}.
  
\methodname{DAGNN} (Domain Adversarial GNN)~\cite{wu2019domain} is a method motivated by DANN~\cite{ganin2016domain} that is one OOD generalization algorithm to learn domain (or environment) invariant graph representations by advocating domain-adversarial learning between the domain classifier and the encoder.
In particular, the first objective is to minimize the classification loss in terms of the encoder on the source domain data, and the second objective aims to facilitate the differentiation between the source and target domains. Such graph adversarial training strategy can maximally utilize the domain information to train classifiers for OOD generalized predictions classification. Note that this method is proposed for text classification where the graphs are converted from the documents, thus the domain (or environment) splits are available in the dataset.

\methodname{GNN-DRO}~\cite{sadeghi2021distributionally} adopts distributionally robust optimization~\cite{rahimian2019distributionally} that is one type of classical algorithm to solve the OOD generalization problem for node-level tasks. 
The GNN model is trained by minimizing the worst expected loss over the considered Wasserstein ball, following the assumption that the data distribution resides in a Wasserstein ball centered at empirical data distribution.

In addition to directly extending existing OOD approaches for general machine learning to graph data as discussed above, there are some other works taking more account of the properties of graph itself.

\methodname{GraphAT} (Graph Adversarial Training)~\cite{feng2019graph} aims to improve the model's generalization via exploring the adversarial training on graphs.
When generating adversarial perturbations on a target sample, GraphAT maximizes the divergence between the prediction of the target sample and its connected samples, meaning that the adversarial perturbations should affect the graph smoothness as much as possible.
After that, GraphAT minimizes the graph adversarial regularizer to update model parameters, reducing the divergence between the prediction of the perturbed target sample and its connected samples.
And a linear approximation method for calculating the adversarial perturbations efficiently is derived based on back-propagation.
By resisting the worst-case perturbations, it can enhance model robustness and generalization.

\methodname{CAP} (Co-Adversarial Perturbation)~\cite{xue2021cap} is proposed from the perspective of loss landscapes during training process. The authors observe GNNs are prone to falling into sharp local minima in loss landscapes in terms of model weight and feature. Therefore, they propose co-adversarial perturbation (CAP) optimization to flatten the weight and feature loss landscapes alternately, which can avoid falling into locally sharp minima and improve generalization ability.
Typically, they formulate the co-adversarial training objective to minimize the maximum training loss within a couple regions of model weights and node features.
For further tackling the efficiency problem of co-adversarial training, they decouple the training objective and devise the alternating adversarial perturbations: one step to conduct the adversarial weight perturbation and training GNNs, as well as another step to calculate the adversarial feature perturbation for each node to update GNNs. 

\methodname{WT-AWP} (Weighted Truncated Adversarial Weight Perturbation)~\cite{wu2022adversarial} follows the line that flatting local minima to improve generalization for OOD graph data.
Since directly applying existing adversarial weight perturbation techniques to train GNNs is not effective in practice induced by the vanishing-gradient issue, WT-AWP uses the loss of adversarial weight perturbation as an additional regularizer with the loss function (e.g., standard cross-entropy) for training GNN. 
It also proposes to remove perturbation in the last layer of the GNN for a more fine-grained control of the training dynamics.
Besides the practical designs for training strategy, a generalization bound for OOD graph classification tasks is also derived.

\methodname{OAD} (Online Adversarial Distillation)~\cite{wang2021online} is an online adversarial knowledge distillation technique for GNNs. 
Different from the above methods that introduce adversarial training into the training process of GNNs, this method brings adversarial training to solve the problem caused by the knowledge distillation. 
Motivated by the knowledge distillation technique can improve the OOD generalization, OAD proposes to train a group of student GNNs in an online fashion with both global and local knowledge.
By transferring informative knowledge of teacher network, the OOD generalization performance of student network can be enhanced.
To learn the complex structure of the local knowledge, adversarial cyclic learning is proposed to achieve more accurate embedding alignment among student models.
OAD method is not only more efficient than vanilla knowledge distillation technique with fewer parameters, but also more effective to handle the graph distribution shift problem.

\subsection{Graph Self-supervised Learning}\label{sec:ssl}

Finally, we introduce the graph self-supervised learning methods for OOD generalization.

Self-supervision as an emerging technique has been employed to train neural networks for more generalizable predictions on the image field~\cite{dou2019domain, mahajan2021domain, zhang2022correct}. 
It is also shown that self-supervised learning can benefit GNNs in gaining more generalization ability~\cite{you2020does}, whose motivations are as follows.
First, the self-supervised learning tasks encourage the GNN models to capture salient critical information of the input graph while avoiding the learned representations trivially overfitting ``shortcuts'' information as supervised learning, leading to better OOD generalization.
Then, Xu \emph{et al.}~\shortcite{xu2021neural} also attribute such success to that self-supervised learning could map semantically similar data to similar representations and therefore some OOD testing data might fall inside the training distribution after the mapping. 

Here we mainly review the typical graph self-supervised methods that claim to improve the graph OOD generalization. For more details of other graph self-supervised methods, the readers could refer to the surveys~\cite{liu2021graph, xie2022self}.

\methodname{Pretraining-GNN}~\cite{hu2019strategies} explores several graph pre-training techniques on both node-level and graph-level to improve OOD generalization of GNNs.
They encourage GNNs to capture domain-specific knowledge about nodes and edges, in addition to graph-level knowledge such that the learned representations can be more OOD generalized.
For node-level pre-training of GNNs, they propose two self-supervised methods, i.e., context prediction and attribute masking.
For graph-level pre-training of GNNs, they also provide two options including making predictions about domain-specific attributes of entire graphs (e.g., supervised labels), or making predictions about graph structure namely modeling the structural similarity of two graphs.
Overall, such pre-training strategy for GNNs is to first perform node-level self-supervised pre-training and then graph-level multi-task supervised pre-training.

\methodname{PATTERN}~\shortcite{yehudai2021local} is proposed to study the ability of GNNs to generalize from small to large graphs, by proposing a self-supervised pretext task that aims at learning useful $d$-pattern representations.
Although GNNs can naturally be applied to graphs with different sizes, it is largely unknown about the mechanism of such size OOD generalization of GNNs.
Therefore, the authors first formalize a representation of local structures called $d$-patterns for characterizing generalization to new graph sizes. 
The $d$-patterns generalize the notion of node degrees to a $d$-step neighborhood of the center node, which models the values of the node and its $d$-step neighbors, as seen by GNNs.
It is proved that even only a small discrepancy in the $d$-patterns distribution between the testing and training
distributions may result in weight assignments that do not
generalize well, indicating the existence of bad global minima with poor generalization.
Then, the self-supervised pretext task is proposed aiming at learning useful $d$-patterns representations from both small and large graphs improving the OOD generalization on graph size with noticeable gains.

\methodname{DR-GST} (Distribution Recovered Graph Self-Training)~\cite{liu2022confidence} is a graph self-training framework that can recover the original labeled dataset without distribution shifts.
Specifically, it first shows that the equality of loss function in self-training framework under the distribution shifts and the population distribution if each pseudo-labeled node is weighted by a proper coefficient.  
Due to the intractability of the coefficient, it replaces the coefficient with the information gain after discovering the same changing trend between them. The information gain is respectively estimated via both dropout variational inference and dropedge variational inference. 
Then, it can recover the shifted distribution with the proposed information gain weighted loss function, which forces the GNN to focus on nodes with high information gain. 
Overall, DR-GST tackles the distribution shift problem from the perspective of information gain, and proposes a loss correction strategy to improve qualities of pseudo labels.
Therefore, more unlabeled nodes can be assigned with pseudo labels whose distribution is the same as that of labeled nodes so as to benefit the OOD generalization ability.

\begin{table*}[ht!]
\centering
\small
\caption{Commonly used synthetic and real-world graph datasets for OOD generalization. ``Task'' denotes each dataset can be used in graph-level or node-level task. ``Type'' indicates what kind of graph data that each dataset includes. ``Cause of Shifts'' indicates the reason for inducing distribution shifts between training and testing data. ``Metric'' is the evaluation metric adopted by each dataset. And ``References'' denotes the work developing each dataset.
}
\label{table:dataset}
\begin{adjustbox}{max width=\linewidth}
\begin{tabular}{lccccc}
\toprule
\textbf{Dataset} & \textbf{Task} & \textbf{Type}        & \textbf{Cause of Shifts} & \textbf{Metric} & \textbf{References}              \\
\midrule
Spurious-Motif       & Graph         &   Synthetic Graph   &   Correlations      & Accuracy        & \cite{shirley2022dir} \\
MNIST-75sp       & Graph         & Superpixel Graph     & Feature Noises           & Accuracy        & \cite{knyazev2019understanding} \\
CMNIST-75sp       & Graph         & Superpixel Graph     & Feature Colors           & Accuracy        & \cite{ding2021closer, gui2022good} \\
D$\&$D$_{200}$   & Graph         & Molecular Graph      & Graph Size               & Accuracy        & \cite{knyazev2019understanding} \\
Graph-SST2       & Graph         & Text Sentiment       & Node Degree              & Accuracy        & \cite{yuan2022explainability}           \\
OGBG-Molhiv      & Graph         & Molecular Graph       & Scaffold                 & ROC-AUC         & \cite{hu2020open}               \\
OGBG-Molpcba     & Graph         & Molecular Graph       & Scaffold                 & Average Precision              & \cite{hu2020open}               \\
OGBG-PPA         & Graph         & Protein Network      & Species                  & Accuracy        & \cite{hu2020open}                       \\
DrugOOD          & Graph         & Molecular Graph      & Assay/Scaffold/Size                  & Accuracy/AUC        & \cite{ji2022drugood}                       \\
\midrule
CBA-Shapes      & Node          & Synthetic Graph       & Feature Colors                & Accuracy        & \cite{gui2022good}            \\
Facebook-100     & Node          & Social Network       & Structure                & Accuracy        & \cite{wu2022towards}            \\
WebKB     & Node          & Webpage Network       & Structure                & Accuracy        & \cite{gui2022good}            \\
Twitch-Explicit  & Node          & Social Network       & Structure                & ROC-AUC         & \cite{rozemberczki2021multi}    \\
Elliptic         & Node          & Bitcoin Transactions & Time                     & F1 Score        & \cite{pareja2020evolvegcn}            \\
OGBN-Arxiv       & Node          & Citation Network     & Time                     & Accuracy        & \cite{hu2020open}               \\
OGBN-Proteins    & Node          & Protein Network      & Species                  & ROC-AUC         & \cite{hu2020open}               \\
OGBN-Products    & Node          & Co-purchasing        & Popularity               & Accuracy        & \cite{hu2020open}               \\
\bottomrule
\end{tabular}
\end{adjustbox}
\end{table*}

Besides, graph contrastive learning can also be adopted to promote OOD generalization. 

\methodname{GraphCL} (Graph Contrastive Learning)~\cite{you2020graph} is one of the representative self-supervised learning methods for GNNs and has shown its generalization ability in practice. 
The authors argue that self-supervision with handcrafted pretext tasks relies on heuristics to design, and thus could limit the generality of the learned graph representations. Therefore, they develop the contrastive learning method GraphCL, whose key idea is to make graph representations agree with each other under the proposed four types of transformations for the input graph.
The generalizability ability of GraphCL is verified on molecular property prediction in chemistry and protein function prediction in biology.

\methodname{RGCL} (Rationale-aware Graph Contrastive Learning)~\cite{li2022let} is proposed to automatically discover rationales as graph augmentations in contrastive learning for further improving the generalization performance in unseen domains with distribution shifts. The authors claim that despite promising performance of some representative methods like GraphCL, etc., the intrinsic random nature makes them suffer from potential semantic information loss, thus hardly capturing the salient information and undermining the generality ability.
RGCL is proposed to tackle this problem, which consists of two modules, i.e., rationale generator and contrastive learner. The rationale generator decides fractions to reveal and conceal in the graph, and yields the rationale encapsulating its instance-discriminative information. The contrastive learner makes use of rationale-aware views to perform instance-discrimination of graphs. Thus, RGCL can prevent losing the discriminative semantics in augmented views as random augmentation and in turn preserve more rationale information with great generalization ability.

\methodname{GAPGC} (Graph Adversarial Pseudo Group Contrast)~\cite{chen2022graphtta} is a test-time training method designed for GNNs with a contrastive loss variant as the self-supervised objective during testing. 
Recently the effectiveness of test-time training has been validated to improve the performance on OOD test data, where some self-supervised auxiliary tasks are proposed. 
The authors argue that the simple augmentations in self-supervised training (e.g., randomly dropping nodes or edges) could harm the label-related critical information in graph representations.
Therefore, GAPGC generates relatively reliable pseudo-labels, avoiding the severe shifts caused by the incorrect positive samples.
The proposed adversarial learnable augmenter and group pseudo-positive samples can promote the relevance between the self-supervised task and the main task, so as to enhance the performance of the main task.  
The theoretical evidence is also derived to show that GAPGC can capture minimal sufficient information for the main task from information theory perspective, which benefits the predictions on the OOD testing data.

\methodname{GT3} (Graph Test-Time Training with Constraint)~\cite{wang2022test} is another test-time training method on graphs, which proposes a hierarchical self-supervised learning framework.
Specifically, it first introduces the global contrastive learning strategy to encourage node representations to capture the global information of the whole graph. The global contrastive learning is based on maximizing the mutual information between the local node representation and the global graph representation.
Then, it presents the local contrastive learning for distinguishing different nodes from different augmented views of a graph, so that the node representation can capture more local information.
Besides, an additional constraint is proposed to encourage that the representations of testing samples are close to the representations of the training samples. The model's OOD generalization capacity 
for the graph classification task can be enhanced based on this  test time training strategy with self-supervised learning.

\section{Theory}\label{sec-theory}
In this section, we review some literature focusing on theoretical analyses of the generalization of GNNs, which are mainly developed to derive the generalization bound of GNNs based on different  statistical learning theories.

\authorname{Scarselli \emph{et al.}}~\cite{scarselli2018vapnik} provide a generalization bound for GNNs based on VC-dimension~\cite{vapnik2015uniform}. 
The authors find that the upper bounds on the VC-dimension for GNNs are comparable to the upper bounds for the recurrent neural networks, meaning that the generalization capability of GNNs increases with the number of connected nodes.
\authorname{Verma \& Zhang}~\cite{verma2019stability} take a further step towards deriving a theoretical analysis of graph convolutional networks (GCNs)~\cite{kipf2016semi} based on algorithmic stability~\cite{bousquet2002stability} and provide generalization bounds for one-layer GCNs. 
They conclude that one-layer GCNs with stable graph convolution filters can satisfy the strong notion of uniform stability and therefore are generalizable. 

\authorname{Garg \emph{et al.}}~\cite{garg2020generalization} study the generalization properties of GNNs on graph classification based on Rademacher complexity.
The generalization analysis explicitly considers the local permutation invariance of the GNN aggregation function.
The derived Rademacher bounds are tighter than the VC bounds from ~\cite{scarselli2018vapnik} for GNNs.
\authorname{Lv}~\cite{lv2021generalization} adopts similar theoretical basis with the work~\cite{garg2020generalization}, providing the Rademacher complexity bound for GCNs with one single hidden layer.  The primary difference is that this work accounts for the specific node-level task of GCNs, which only involves a fixed adjacency matrix.

\authorname{Liao \emph{et al.}}~\cite{liao2020pac} establish a PAC-Bayesian generalization bound of GNNs on graph classification. It further improves upon the Rademacher complexity based bound proposed in the work~\cite{garg2020generalization}, deriving a tighter dependency on the maximum node degree and the maximum hidden dimension. Also, \authorname{Ma \emph{et al.}}~\cite{ma2021subgroup} present a PAC-Bayesian analysis for generalization performances of GNNs on subgroups of nodes under non-IID node-level tasks, which is the key difference compared with the work~\cite{liao2020pac}.

\authorname{Du \emph{et al.}}~\cite{du2019graph} establish Graph Neural Tangent Kernel (GNTK) to characterize the generalization bound of GNNs on graph classification. Note that GNTK is induced by infinitely wide GNNs, whose prediction depends only on pairwise kernel values between graphs, and can be calculated efficiently with an analytic formula. It enjoys the expressive power of GNNs, while inheriting the benefits of graph kernels, e.g., easy to train, provable theoretical guarantees, etc.
Relying on the GNTK method, \authorname{Xu \emph{et al.}}~\cite{xu2021neural} derive theoretical evidence of generalization capabilities in one-layer GNNs and study the effect of the alignment of network architecture and target algorithmic tasks on OOD generalization. 
Along with this line, \authorname{Zhang \emph{et al.}}~\cite{zhang2020fast} prove that using proper tensor initialization and accelerated gradient descent, their algorithm can learn a GNN with one hidden layer having the zero generalization error for regression problems or sufficiently close to the ground-truth model, assuming such a ground-truth model exists.

Considering most methods mentioned above are developed based on that graph data can be generated and labeled in any arbitrary way which is hard to be satisfied in practice, some works establish generalization bounds that depend on the graph data as follows.

\methodname{Baranwal \emph{et al.}}~\cite{baranwal2021graph} study OOD generalization of GNNs under a specific data generating mechanism namely contextual stochastic block model and analyze the relation between linear separability and OOD generalization on graphs. The generalization guarantee for one-layer GCNs on binary node classification is derived. 
Furthermore, \authorname{Maskey \emph{et al.}}~\cite{maskey2022generalization} consider a generative model graphons for the graphs which is not only theoretically powerful and general, but allows much tighter generalization bounds.

\section{Datasets for Evaluation}\label{sec-datasets}

To promote further research of graph OOD generalization, we summarize the existing popular graph datasets for evaluation in Table~\ref{table:dataset}.
There are two groups of datasets, where one group is for graph-level tasks and the other group is for node-level tasks.
These datasets cover multiple sources of graphs (e.g., social network, citation network, molecular graph, etc) and their causes of distribution shifts are also complex and diverse (e.g., time, species, scaffold, etc.).

\subsection{Datasets for Graph-level Tasks}
First, we review some representative datasets for evaluating the model performances on graph classification tasks. 

\methodname{Spurious-Motif}~\cite{shirley2022dir}: It is a synthetic dataset created by following the work~\cite{ying2019gnnexplainer}, which is designed for distribution shifts on graph structure. Each graph consists of one motif and one base subgraph.
The base subgraph includes Tree, Ladder, and Wheel (denoted by $V = 0, 1, 2$, respectively) and the motif includes Cycle, House, and Crane (denoted by $I = 0, 1, 2$). 
The ground-truth label $Y$ only depends on the motif $I$, which is sampled uniformly. 
The spurious correlation between $V$ and $Y$ is injected by controlling the base subgraphs distribution as: $P(V) = b$ if $V = I$ and $P(V) = (1-b)/2$ if $V=I$.
Intuitively, $b$ controls the strength of the spurious correlation. It can set $b$ to different values in the testing and training set to simulate the distribution shifts.

\methodname{MNIST-75sp}~\cite{knyazev2019understanding}: It is a semi-artificial dataset, where each graph is converted from an image in MNIST~\cite{lecun1998gradient} using superpixels~\cite{achanta2012slic}. The nodes are superpixels, and the edges are calculated by the spatial distance between nodes. The node features are set as the super-pixel coordinates and intensity. The task is to classify each graph into the corresponding handwritten digit labeled from $0$ to $9$. To simulate distribution shifts with respect to graph features, it generates testing graphs by colorizing images, i.e., adding two more channels and adding independent Gaussian noise to each channel.

\methodname{CMNIST-75sp}~\cite{ding2021closer, gui2022good}: It is also a semi-artificial dataset, consisting of graphs converted from the images in MNIST using superpixels. Different from MNIST-75sp that adds noise to simulate distribution shifts, CMNIST-75sp colorizes the digits with different colors according to the digit labels or dataset split, inspired by the work~\cite{arjovsky2019invariant}. Note that there are two choices of CMNIST-75sp to simulate the covariate shifts or concept shifts respectively. For the former choice, the testing data are colorized with unseen colors compared with the colors for the training data. For the latter choice, the colors are correlated with the digit labels for the training data, while colors have different correlations with labels for testing data, respectively.

\methodname{D$\&$D$_{200}$}~\cite{knyazev2019understanding}: It is a real-world graph classification dataset that consists of 1,178 protein network structures with 82 discrete node labels.
The task is to classify each graph into enzyme or non-enzyme class.
To create distribution shifts on graph sizes, the training and testing sets are split by graph sizes, i.e., the models are trained on small graphs but tested on larger graphs.
Specifically,
the training set includes graphs with 30 to 200 nodes while the testing set includes graphs with 201 to 5,748 nodes.

\methodname{Graph-SST2}~\cite{yuan2022explainability}: 
It is a real-world graph dataset originating from a natural language sentimental analysis dataset.
Each graph is converted from a text sequence, where nodes represent words, edges indicate relations between words, and label is the sentence sentiment. Graphs are split into different sets according to their average node degree to create distribution shifts. 
The node features are initialized by the pre-trained BERT word embedding~\cite{devlin2018bert}. 
Thanks to the semantics of these graphs, this dataset is more human-understandable for visualizing or analyzing some intermediate results.

\methodname{OGBG}~\cite{hu2020open}: 
Open Graph Benchmark (OGB) is a benchmark consisting of realistic, large-scale, and diverse datasets for machine learning on graphs, where OGBG is a subset including several representative datasets for evaluation OOD generalization in graph-level tasks, e.g., OGBG-Molhiv, OGBG-Molpcba, OGBG-PPA, etc.
Specifically, \methodname{OGBG-Molhiv} and \methodname{OGBG-Molpcba} are two graph property prediction datasets with distribution shifts. The task is to predict the target molecular properties. The dataset provides the default scaffold splitting procedure, i.e., splitting the graphs based on their two-dimensional structural frameworks. Note that this scaffold splitting strategy aims to separate structurally different molecules into different subsets, which provides a more realistic and challenging scenario for testing graph OOD generalization. And \methodname{OGBG-PPA} consists of undirected protein association neighborhoods extracted from the protein-protein association networks of 1,581 different species. The task is to predict what taxonomic group the given protein association neighborhood graph originates from. The dataset adopts species split, i.e., separating graphs from different species into different subsets. 

\methodname{DrugOOD}~\cite{ji2022drugood}: It is a benchmark for AI-aided drug discovery, including some realistic molecular graph datasets. 
It provides an automated pipeline for curating OOD datasets based on a large-scale bioassay dataset ChEMBL~\cite{mendez2019chembl}.
Also, it presents more diverse dataset splitting indicators than OGB to generate specific domains that are aligned with the domain knowledge of biochemistry. Rather than only adopting scaffold as the indicator of dataset splitting, it can provide more choices for separating graphs into different subsets in terms of assay and size to create distribution shifts.

\subsection{Datasets for Node-level Tasks}
Then, we review some representative datasets for evaluating the model performances on node classification tasks.

\methodname{CBA-Shapes}~\cite{gui2022good}: It is a synthetic dataset created by following the BA-Shapes dataset from the work~\cite{ying2019gnnexplainer}.
The input graph contains a base graph and a set of motifs, where the base graph is a Barab\'{a}si-Albert (BA) graph on 300 nodes and the set of motifs includes 80 house-structured motifs.
The task is to predict the structural role of each node, including the top, middle, or bottom node of a house-structured motif, or the node from the base graph, i.e., a 4-class classification task. 
Node features are assigned with colors to create distribution shifts, which also have two choices to simulate the covariate shifts or concept shifts. For the former choice, the testing nodes are colorized with unseen colors compared with the colors of the training nodes. For the latter choice, the colors are correlated with the labels of the training nodes, while colors have different correlations with labels of the testing nodes, respectively.

\methodname{Facebook-100}~\cite{wu2022towards}:  It is a real-world node classification dataset which consists of 100 Facebook social network snapshots from the year 2005.
Each network contains nodes as Facebook users from a specific American university. The distribution shifts can be introduced by splitting training and testing sets via selecting different universities that the users in a network are from, since these networks have significantly diverse sizes, densities and degree distributions. For example, the default dataset split in the work~\cite{wu2022towards} is to adopt the corresponding networks from three of fourteen universities (e.g., John Hopkins, Cornell, etc.) as training set, and the network from another three universities (i.e., Penn, Brown and Texas) as the testing set. Of course, the other combinations can also be used to evaluate the node-level OOD generalization ability.

\methodname{WebKB}~\cite{gui2022good}: It is a real-world university webpage network dataset for node classification.
The nodes denote webpages and edges are hyperlinks between two webpages. The node features are from the words appearing in the webpage. The task is to predict the classes of webpages including student, project, course, staff, or faculty.
The distribution shifts are from splitting the dataset conforming to the domain university. Therefore, the OOD generalized predictions can be achieved when only using the word contents and hyperlinks of webpages rather than using the university features.

\methodname{Twitch-Explicit}~\cite{rozemberczki2021multi}: It is a real-world social network dataset, where nodes are Twitch users and edges are friendships between two users.  Node features are games liked, location and streaming habits. Each network is collected from a specific region, including DE, ENGB, ES, FR, PTBR, RU and TW. The seven networks have significantly different structural properties, e.g., densities and maximum node degrees~\cite{wu2022towards}. The distribution shifts between training and testing sets are from splitting the dataset according to the network region.

\methodname{Elliptic}~\cite{pareja2020evolvegcn}: It is a realistic Bitcoin transaction network dataset consisting of several snapshots, where nodes are transactions and edges are payment flows. The task is to distinguish between licit and illicit transactions in future data. By adopting older snapshots in terms of time as the training set while newer snapshots as the testing set, the distribution shifts can be observed due to some emerging events in the market.

\methodname{OGBN}~\cite{hu2020open}: It includes some node properties prediction datasets, e.g., OGBN-Arxiv, OGBN-Proteins, and OGBN-Products, which is another subset of the whole OGB~\cite{hu2020open}.
Specifically, \methodname{OGBN-Arxiv} is a real-world citation dataset, where nodes are arXiv papers, and edges are citations between papers. Its 40-class prediction task is to predict the subject area of arXiv papers. The node distribution shifts are introduced by splitting papers from different time ranges into training and testing sets.
And \methodname{OGBN-Proteins} a protein graph, where nodes represent proteins and edges indicate different types of biologically meaningful associations between proteins. The task is to predict the presence of protein functions. The distribution shifts are introduced by splitting the protein nodes into different subsets according to the species that the proteins come from. In addition, \methodname{OGBN-Products} is an Amazon product co-purchasing network. Nodes represent products in Amazon, and edges indicate that the two products are purchased together. The task is to predict the product category. The distribution shifts are created by a more challenging and realistic dataset splitting according to the popularity of products, i.e., using the popular products for training but relatively unpopular products for testing.

\subsection{Other Benchmarks}
In addition, there are also some works that collect these commonly used or more than one datasets above into a standard evaluation open-source benchmark and report the experimental results for some well-known general OOD algorithms and graph OOD methods under the proposed evaluation protocols. 
Since the details of most datasets have been discussed above, here we review these packages briefly.
Specifically, \methodname{GDS}~\cite{ding2021closer} collects eight datasets for graph-level tasks reflecting a diverse range of distribution shifts across graphs to compare the performance of popular OOD generalization algorithms and GNN backbones. 
\methodname{GOOD}~\cite{gui2022good} summarizes more than ten  datasets for both graph-level and node-level tasks with diverse types of distribution shifts introduced by combining different domain selection strategies and distribution shift types. It also contains the experiments to show the significant performance gaps between in-distribution and OOD settings and the comparisons among  different OOD methods for both general machine learning and the graph field.

\section{Discussions}\label{sec-discussions}
In this section, we briefly summarize this survey and further discuss several challenges as well as opportunities worthy of future explorations.

\subsection{Summary}\label{subsec-summary}

The diversity and quality of training graph data play an important role in OOD generalization of graph machine learning approaches. Several graph data augmentation methods, including structure-wise, feature-wise, and mixed-type methods are developed to achieve good performances with simple yet effective paradigms. 

Another line of works focuses on exploiting new graph models to promote the OOD generalization capability. Compared to graph data augmentation, these models overall enjoy more solid theoretical ground and more graph-specific designs. The disentanglement-based graph models present good motivations while the causality-based graph models are backed by diverse causal inference theories. These tailored graph models also show promising OOD generalization performances in practice.

Recently, there is a rapid development for graph learning strategies, including graph invariant learning, graph adversarial training, and graph self-supervised learning. Compared with the graph models, these methods pay more attention to the learning process, so that they are more flexible to be compatible with different GNN backbones for enhancing OOD generalization. 

To build the theoretical framework of graph generalization, a number of theoretical derivations on generalization bounds are proposed, which benefit the deeper understanding of graph OOD generalization methods. 
And to promote deeper research, diverse datasets under complex realistic distribution shifts covering node-level and graph-level tasks are adopted to verify the effectiveness of graph OOD generalization methods comprehensively and fairly.

\subsection{Future Directions}\label{subsec-future}

There exist plenty of challenges and opportunities worthy of future explorations.

\subsubsection{More Theoretical Guarantees}

While some graph OOD generalization methods have made great progress empirically, there is still a large gap between these methods and the theories introduced in Section~\ref{sec-theory}. It is a critical step to derive theoretical characterization on a learnable graph OOD generalization problem and further develop methods with theoretical guarantees for OOD optimality. 
Besides, it is also worth figuring out what kind of distribution shifts (e.g., covariate shifts, concept shifts, or even label shifts) and investigating OOD generalization theories built upon the specific assumptions on distribution shifts.

\subsubsection{GNN Architecture}

Recently, some studies~\cite{knyazev2019understanding,velivckovic2019neural,xu2021neural,ko2021learning,customizedgnn} highlight the importance of careful design for GNN architecture (e.g., readout operation) to gracefully generalize to OOD graph data. 
Besides hand-crafted model designs, automatically tailoring a customized GNN architecture suitable for each graph instance also benefits the predictions under distribution shifts~\cite{qin2022graph}.
It remains to be further explored how to design theoretically guaranteed GNN architectures for OOD generalization. And more research efforts need to be paid on automatically learning OOD generalized GNN architectures suitable for diverse environments.

\subsubsection{Environment Split}
The majority of general OOD generalization algorithms require multiple training environments~\cite{shen2021towards}. 
However, it is prohibitively expensive to collect accurate environment labels for real-world graphs, limiting the adoption of those algorithms. 
It is worth investigating to develop the single environment graph OOD generalization method or infer environment split accurately during training.
Moreover, for many real-world situations, graph data often changes/evolves over time, which resides in dynamic or even continuous environments~\cite{galke2021lifelong}. 
So it remains a promising future direction to perform graph OOD generalization dynamically or continuously that efficiently updates graph models or learning strategies in terms of time to generalize to new data under unknown distribution.

\subsubsection{Test-Time Training for Generalization}
Graph test-time training can allow more flexibility in inference time to make use of the inference unlabeled data during the testing stage. 
It can improve the graph OOD generalization under unknown distribution shifts via solving a test-time task.
In addition to the two works~\cite{chen2022graphtta, wang2022test} introduced in Section~\ref{sec:ssl} that adopt contrastive test-time tasks, one more recent attempt \methodname{GTrans}~\cite{jin2022empowering} proposes to adapt and refine graph data at test-time.
It is a valuable direction to design more test-time training tasks or explore more test-time training strategies to improve OOD generalization on graphs.

\subsubsection{Broader Scope of Applications}
OOD graph data widely exist in our daily life.
Although some classical machine learning approaches on graphs have been utilized on various realistic applications, it is a promising direction to deploy the OOD generalized graph methods in the wild where distribution shifts widely exist~\cite{han2021reliable, yang2022learning, sinha2020evaluating, li2021does, li2022critical}, including recommender systems, social networks, traffic prediction, materials science, and risk-sensitive finance or healthcare fields, for more effective, trustworthy and satisfying predictions. How to incorporate proper domain knowledge is one main challenge to apply graph OOD generalization into these applications.
One possible principle is to treat the integrated domain knowledge as additional prior knowledge to guide the designs of graph models and learning strategies.

\bibliographystyle{IEEEtran}
\small
\bibliography{reference}

\begin{thebibliography}{100}
\providecommand{\url}[1]{#1}
\csname url@samestyle\endcsname
\providecommand{\newblock}{\relax}
\providecommand{\bibinfo}[2]{#2}
\providecommand{\BIBentrySTDinterwordspacing}{\spaceskip=0pt\relax}
\providecommand{\BIBentryALTinterwordstretchfactor}{4}
\providecommand{\BIBentryALTinterwordspacing}{\spaceskip=\fontdimen2\font plus
\BIBentryALTinterwordstretchfactor\fontdimen3\font minus
  \fontdimen4\font\relax}
\providecommand{\BIBforeignlanguage}[2]{{%
\expandafter\ifx\csname l@#1\endcsname\relax
\typeout{** WARNING: IEEEtran.bst: No hyphenation pattern has been}%
\typeout{** loaded for the language `#1'. Using the pattern for}%
\typeout{** the default language instead.}%
\else
\language=\csname l@#1\endcsname
\fi
#2}}
\providecommand{\BIBdecl}{\relax}
\BIBdecl

\bibitem{qiu2018deepinf}
J.~Qiu, J.~Tang, H.~Ma, Y.~Dong, K.~Wang, and J.~Tang, ``Deepinf: Social
  influence prediction with deep learning,'' in \emph{Proceedings of the 24th
  ACM SIGKDD International Conference on Knowledge Discovery \& Data Mining},
  2018, pp. 2110--2119.

\bibitem{wu2020graph}
S.~Wu, F.~Sun, W.~Zhang, X.~Xie, and B.~Cui, ``Graph neural networks in
  recommender systems: a survey,'' \emph{ACM Computing Surveys}, vol.~55,
  no.~5, pp. 1--37, 2022.

\bibitem{wang2017knowledge}
Q.~Wang, Z.~Mao, B.~Wang, and L.~Guo, ``Knowledge graph embedding: A survey of
  approaches and applications,'' \emph{IEEE Transactions on Knowledge and Data
  Engineering}, 2017.

\bibitem{yu2018spatio}
B.~Yu, H.~Yin, and Z.~Zhu, ``Spatio-temporal graph convolutional networks: a
  deep learning framework for traffic forecasting,'' in \emph{International
  Joint Conference on Artificial Intelligence}, 2018.

\bibitem{bengio2019meta}
Y.~Bengio, T.~Deleu, N.~Rahaman, R.~Ke, S.~Lachapelle, O.~Bilaniuk, A.~Goyal,
  and C.~Pal, ``A meta-transfer objective for learning to disentangle causal
  mechanisms,'' \emph{International Conference on Learning Representations},
  2019.

\bibitem{li2021ood}
H.~Li, X.~Wang, Z.~Zhang, and W.~Zhu, ``Ood-gnn: Out-of-distribution
  generalized graph neural network,'' \emph{IEEE Transactions on Knowledge and
  Data Engineering}, 2022.

\bibitem{hu2020open}
W.~Hu, M.~Fey, M.~Zitnik, Y.~Dong, H.~Ren, B.~Liu, M.~Catasta, and J.~Leskovec,
  ``Open graph benchmark: Datasets for machine learning on graphs,''
  \emph{Neural Information Processing Systems (NeurIPS)}, 2020.

\bibitem{yang2020financial}
S.~Yang, Z.~Zhang, J.~Zhou, Y.~Wang, W.~Sun, X.~Zhong, Y.~Fang, Q.~Yu, and
  Y.~Qi, ``Financial risk analysis for smes with graph-based supply chain
  mining.'' in \emph{International Joint Conference on Artificial
  Intelligence}, 2020.

\bibitem{agarwal2021towards}
C.~Agarwal, H.~Lakkaraju, and M.~Zitnik, ``Towards a unified framework for fair
  and stable graph representation learning,'' in \emph{UAI}, 2021.

\bibitem{liang2020learning}
M.~Liang, B.~Yang, R.~Hu, Y.~Chen, R.~Liao, S.~Feng, and R.~Urtasun, ``Learning
  lane graph representations for motion forecasting,'' in \emph{European
  Conference on Computer Vision}.\hskip 1em plus 0.5em minus 0.4em\relax
  Springer, 2020, pp. 541--556.

\bibitem{shlomi2020graph}
J.~Shlomi, P.~Battaglia, and J.-R. Vlimant, ``Graph neural networks in particle
  physics,'' \emph{Machine Learning: Science and Technology}, vol.~2, no.~2, p.
  021001, 2020.

\bibitem{panagopoulos2020transfer}
G.~Panagopoulos, G.~Nikolentzos, M.~Vazirgiannis \emph{et~al.}, ``Transfer
  graph neural networks for pandemic forecasting,'' \emph{Association for the
  Advancement of Artificial Intelligence}, 2021.

\bibitem{horry2020covid}
M.~J. Horry, S.~Chakraborty, M.~Paul, A.~Ulhaq, B.~Pradhan, M.~Saha, and
  N.~Shukla, ``Covid-19 detection through transfer learning using multimodal
  imaging data,'' \emph{Ieee Access}, 2020.

\bibitem{hsieh2021drug}
K.~Hsieh, Y.~Wang, L.~Chen, Z.~Zhao, S.~Savitz, X.~Jiang, J.~Tang, and Y.~Kim,
  ``Drug repurposing for covid-19 using graph neural network and harmonizing
  multiple evidence,'' \emph{Scientific reports}, vol.~11, no.~1, pp. 1--13,
  2021.

\bibitem{shen2021towards}
Z.~Shen, J.~Liu, Y.~He, X.~Zhang, R.~Xu, H.~Yu, and P.~Cui, ``Towards
  out-of-distribution generalization: A survey,'' \emph{arXiv preprint
  arXiv:2108.13624}, 2021.

\bibitem{wang2021generalizing}
J.~Wang, C.~Lan, C.~Liu, Y.~Ouyang, W.~Zeng, and T.~Qin, ``Generalizing to
  unseen domains: A survey on domain generalization,'' \emph{International
  Joint Conference on Artificial Intelligence}, 2021.

\bibitem{zhou2022domain}
K.~Zhou, Z.~Liu, Y.~Qiao, T.~Xiang, and C.~C. Loy, ``Domain generalization: A
  survey,'' \emph{IEEE Transactions on Pattern Analysis and Machine
  Intelligence}, 2022.

\bibitem{ding2022data}
K.~Ding, Z.~Xu, H.~Tong, and H.~Liu, ``Data augmentation for deep graph
  learning: A survey,'' \emph{arXiv preprint arXiv:2202.08235}, 2022.

\bibitem{zhao2022graph}
T.~Zhao, G.~Liu, S.~G{\"u}nnemann, and M.~Jiang, ``Graph data augmentation for
  graph machine learning: A survey,'' \emph{arXiv preprint arXiv:2202.08871},
  2022.

\bibitem{liu2021graph}
Y.~Liu, M.~Jin, S.~Pan, C.~Zhou, Y.~Zheng, F.~Xia, and P.~Yu, ``Graph
  self-supervised learning: A survey,'' \emph{IEEE Transactions on Knowledge
  and Data Engineering}, 2022.

\bibitem{xie2022self}
Y.~Xie, Z.~Xu, J.~Zhang, Z.~Wang, and S.~Ji, ``Self-supervised learning of
  graph neural networks: A unified review,'' \emph{IEEE Transactions on Pattern
  Analysis and Machine Intelligence}, 2022.

\bibitem{sun2018adversarial}
L.~Sun, Y.~Dou, C.~Yang, K.~Zhang, J.~Wang, S.~Y. Philip, L.~He, and B.~Li,
  ``Adversarial attack and defense on graph data: A survey,'' \emph{IEEE
  Transactions on Knowledge and Data Engineering}, 2022.

\bibitem{chen2020survey}
L.~Chen, J.~Li, J.~Peng, T.~Xie, Z.~Cao, K.~Xu, X.~He, and Z.~Zheng, ``A survey
  of adversarial learning on graphs,'' \emph{arXiv preprint arXiv:2003.05730},
  2020.

\bibitem{zhao2020data}
T.~Zhao, Y.~Liu, L.~Neves, O.~Woodford, M.~Jiang, and N.~Shah, ``Data
  augmentation for graph neural networks,'' \emph{Association for the
  Advancement of Artificial Intelligence}, 2021.

\bibitem{park2021metropolis}
H.~Park, S.~Lee, S.~Kim, J.~Park, J.~Jeong, K.-M. Kim, J.-W. Ha, and H.~J. Kim,
  ``Metropolis-hastings data augmentation for graph neural networks,''
  \emph{Advances in Neural Information Processing Systems}, vol.~34, 2021.

\bibitem{wuknowledge}
L.~Wu, H.~Lin, Y.~Huang, and S.~Z. Li, ``Knowledge distillation improves graph
  structure augmentation for graph neural networks,'' in \emph{Neural
  Information Processing Systems}, 2022.

\bibitem{feng2020graph}
W.~Feng, J.~Zhang, Y.~Dong, Y.~Han, H.~Luan, Q.~Xu, Q.~Yang, E.~Kharlamov, and
  J.~Tang, ``Graph random neural networks for semi-supervised learning on
  graphs,'' \emph{Advances in neural information processing systems}, vol.~33,
  pp. 22\,092--22\,103, 2020.

\bibitem{kong2020flag}
K.~Kong, G.~Li, M.~Ding, Z.~Wu, C.~Zhu, B.~Ghanem, G.~Taylor, and T.~Goldstein,
  ``Robust optimization as data augmentation for large-scale graphs,'' in
  \emph{Proceedings of the IEEE/CVF Conference on Computer Vision and Pattern
  Recognition}, 2022, pp. 60--69.

\bibitem{liu2022local}
S.~Liu, R.~Ying, H.~Dong, L.~Li, T.~Xu, Y.~Rong, P.~Zhao, J.~Huang, and D.~Wu,
  ``Local augmentation for graph neural networks,'' in \emph{International
  Conference on Machine Learning}.\hskip 1em plus 0.5em minus 0.4em\relax PMLR,
  2022, pp. 14\,054--14\,072.

\bibitem{you2020graph}
Y.~You, T.~Chen, Y.~Sui, T.~Chen, Z.~Wang, and Y.~Shen, ``Graph contrastive
  learning with augmentations,'' \emph{Advances in Neural Information
  Processing Systems}, vol.~33, pp. 5812--5823, 2020.

\bibitem{liu2022graph}
G.~Liu, T.~Zhao, J.~Xu, T.~Luo, and M.~Jiang, ``Graph rationalization with
  environment-based augmentations,'' in \emph{Proceedings of the 28th ACM
  SIGKDD International Conference on Knowledge Discovery \& Data Mining}, 2022.

\bibitem{yu2022finding}
J.~Yu, J.~Liang, and R.~He, ``Finding diverse and predictable subgraphs for
  graph domain generalization,'' \emph{arXiv preprint arXiv:2206.09345}, 2022.

\bibitem{sui2022adversarial}
Y.~Sui, X.~Wang, J.~Wu, A.~Zhang, and X.~He, ``Adversarial causal augmentation
  for graph covariate shift,'' \emph{arXiv preprint arXiv:2211.02843}, 2022.

\bibitem{zhang2017mixup}
H.~Zhang, M.~Cisse, Y.~N. Dauphin, and D.~Lopez-Paz, ``mixup: Beyond empirical
  risk minimization,'' \emph{International Conference on Learning
  Representations}, 2018.

\bibitem{ma2019disentangled}
J.~Ma, P.~Cui, K.~Kuang, X.~Wang, and W.~Zhu, ``Disentangled graph
  convolutional networks,'' in \emph{International conference on machine
  learning}.\hskip 1em plus 0.5em minus 0.4em\relax PMLR, 2019, pp. 4212--4221.

\bibitem{liu2020independence}
Y.~Liu, X.~Wang, S.~Wu, and Z.~Xiao, ``Independence promoted graph disentangled
  networks,'' in \emph{Association for the Advancement of Artificial
  Intelligence}, vol.~34, no.~04, 2020, pp. 4916--4923.

\bibitem{yang2020factorizable}
Y.~Yang, Z.~Feng, M.~Song, and X.~Wang, ``Factorizable graph convolutional
  networks,'' \emph{Advances in Neural Information Processing Systems},
  vol.~33, pp. 20\,286--20\,296, 2020.

\bibitem{fan2022debiasing}
S.~Fan, X.~Wang, Y.~Mo, C.~Shi, and J.~Tang, ``Debiasing graph neural networks
  via learning disentangled causal substructure,'' in \emph{Advances in Neural
  Information Processing Systems}, 2022.

\bibitem{guo2020interpretable}
X.~Guo, L.~Zhao, Z.~Qin, L.~Wu, A.~Shehu, and Y.~Ye, ``Interpretable deep graph
  generation with node-edge co-disentanglement,'' in \emph{Proceedings of the
  26th ACM SIGKDD international conference on knowledge discovery \& data
  mining}, 2020, pp. 1697--1707.

\bibitem{li2021disentangled}
H.~Li, X.~Wang, Z.~Zhang, Z.~Yuan, H.~Li, and W.~Zhu, ``Disentangled
  contrastive learning on graphs,'' \emph{Advances in Neural Information
  Processing Systems}, vol.~34, 2021.

\bibitem{li2022disentangled}
H.~Li, Z.~Zhang, X.~Wang, and W.~Zhu, ``Disentangled graph contrastive learning
  with independence promotion,'' \emph{IEEE Transactions on Knowledge and Data
  Engineering}, 2022.

\bibitem{fan2021generalizing}
S.~Fan, X.~Wang, C.~Shi, P.~Cui, and B.~Wang, ``Generalizing graph neural
  networks on out-of-distribution graphs,'' \emph{arXiv preprint
  arXiv:2111.10657}, 2021.

\bibitem{fan2022debiased}
S.~Fan, X.~Wang, C.~Shi, K.~Kuang, N.~Liu, and B.~Wang, ``Debiased graph neural
  networks with agnostic label selection bias,'' \emph{IEEE Transactions on
  Neural Networks and Learning Systems}, 2022.

\bibitem{sui2021deconfounded}
Y.~Sui, X.~Wang, J.~Wu, M.~Lin, X.~He, and T.-S. Chua, ``Causal attention for
  interpretable and generalizable graph classification,'' \emph{Proceedings of
  the 28th ACM SIGKDD International Conference on Knowledge Discovery \& Data
  Mining}, 2022.

\bibitem{wu2022deconfounding}
Y.~Wu, X.~Wang, A.~Zhang, X.~Hu, F.~Feng, X.~He, and T.-S. Chua,
  ``Deconfounding to explanation evaluation in graph neural networks,''
  \emph{arXiv preprint arXiv:2201.08802}, 2022.

\bibitem{chen2022learning}
Y.~Chen, Y.~Zhang, Y.~Bian, H.~Yang, K.~Ma, B.~Xie, T.~Liu, B.~Han, and
  J.~Cheng, ``Learning causally invariant representations for
  out-of-distribution generalization on graphs,'' in \emph{Thirty-Sixth
  Conference on Neural Information Processing Systems}, 2022.

\bibitem{bevilacqua2021size}
B.~Bevilacqua, Y.~Zhou, and B.~Ribeiro, ``Size-invariant graph representations
  for graph classification extrapolations,'' in \emph{International Conference
  on Machine Learning}.\hskip 1em plus 0.5em minus 0.4em\relax PMLR, 2021, pp.
  837--851.

\bibitem{zhou2022ood}
Y.~Zhou, G.~Kutyniok, and B.~Ribeiro, ``Ood link prediction generalization
  capabilities of message-passing gnns in larger test graphs,'' \emph{Advances
  in Neural Information Processing Systems}, 2022.

\bibitem{zhao2021counterfactual}
T.~Zhao, G.~Liu, D.~Wang, W.~Yu, and M.~Jiang, ``Learning from counterfactual
  links for link prediction,'' in \emph{International Conference on Machine
  Learning}.\hskip 1em plus 0.5em minus 0.4em\relax PMLR, 2022, pp.
  26\,911--26\,926.

\bibitem{lin2021generative}
W.~Lin, H.~Lan, and B.~Li, ``Generative causal explanations for graph neural
  networks,'' in \emph{International Conference on Machine Learning}.\hskip 1em
  plus 0.5em minus 0.4em\relax PMLR, 2021, pp. 6666--6679.

\bibitem{li2022gil}
H.~Li, Z.~Zhang, X.~Wang, and W.~Zhu, ``Learning invariant graph
  representations under distribution shifts,'' in \emph{Advances in Neural
  Information Processing Systems}, 2022.

\bibitem{shirley2022dir}
Y.-X. Wu, X.~Wang, A.~Zhang, X.~He, and T.~seng Chua, ``Discovering invariant
  rationales for graph neural networks,'' in \emph{International Conference on
  Learning Representations}, 2022.

\bibitem{miao2022interpretable}
S.~Miao, M.~Liu, and P.~Li, ``Interpretable and generalizable graph learning
  via stochastic attention mechanism,'' in \emph{International Conference on
  Machine Learning}, 2022, pp. 15\,524--15\,543.

\bibitem{wu2022towards}
Q.~Wu, H.~Zhang, J.~Yan, and D.~Wipf, ``Handling distribution shifts on graphs:
  An invariance perspective,'' in \emph{International Conference on Learning
  Representations}, 2022.

\bibitem{zhang2022dynamic}
Z.~Zhang, X.~Wang, Z.~Zhang, H.~Li, Z.~Qin, and W.~Zhu, ``Dynamic graph neural
  networks under spatio-temporal distribution shift,'' in \emph{Thirty-Sixth
  Conference on Neural Information Processing Systems}, 2022.

\bibitem{zhu2021shift}
Q.~Zhu, N.~Ponomareva, J.~Han, and B.~Perozzi, ``Shift-robust gnns: Overcoming
  the limitations of localized graph training data,'' \emph{Advances in Neural
  Information Processing Systems}, vol.~34, 2021.

\bibitem{buffelli2022sizeshiftreg}
D.~Buffelli, P.~Li{\`o}, and F.~Vandin, ``Sizeshiftreg: a regularization method
  for improving size-generalization in graph neural networks,'' \emph{Neural
  Information Processing Systems}, 2022.

\bibitem{zhang2021stable}
S.~Zhang, K.~Kuang, J.~Qiu, J.~Yu, Z.~Zhao, H.~Yang, Z.~Zhang, and F.~Wu,
  ``Stable prediction on graphs with agnostic distribution shift,'' \emph{arXiv
  preprint arXiv:2110.03865}, 2021.

\bibitem{wu2019domain}
M.~Wu, S.~Pan, X.~Zhu, C.~Zhou, and L.~Pan, ``Domain-adversarial graph neural
  networks for text classification,'' in \emph{2019 IEEE International
  Conference on Data Mining (ICDM)}.\hskip 1em plus 0.5em minus 0.4em\relax
  IEEE, 2019, pp. 648--657.

\bibitem{sadeghi2021distributionally}
A.~Sadeghi, M.~Ma, B.~Li, and G.~B. Giannakis, ``Distributionally robust
  semi-supervised learning over graphs,'' \emph{arXiv preprint
  arXiv:2110.10582}, 2021.

\bibitem{feng2019graph}
F.~Feng, X.~He, J.~Tang, and T.-S. Chua, ``Graph adversarial training:
  Dynamically regularizing based on graph structure,'' \emph{IEEE Transactions
  on Knowledge and Data Engineering}, vol.~33, no.~6, pp. 2493--2504, 2019.

\bibitem{xue2021cap}
H.~Xue, K.~Zhou, T.~Chen, K.~Guo, X.~Hu, Y.~Chang, and X.~Wang, ``Cap:
  Co-adversarial perturbation on weights and features for improving
  generalization of graph neural networks,'' \emph{arXiv preprint
  arXiv:2110.14855}, 2021.

\bibitem{wu2022adversarial}
Y.~Wu, A.~Bojchevski, and H.~Huang, ``Adversarial weight perturbation improves
  generalization in graph neural networks,'' \emph{Association for the
  Advancement of Artificial Intelligence}, 2023.

\bibitem{wang2021online}
C.~Wang, Z.~Wang, D.~Chen, S.~Zhou, Y.~Feng, and C.~Chen, ``Online adversarial
  distillation for graph neural networks,'' \emph{arXiv preprint
  arXiv:2112.13966}, 2021.

\bibitem{hu2019strategies}
W.~Hu, B.~Liu, J.~Gomes, M.~Zitnik, P.~Liang, V.~Pande, and J.~Leskovec,
  ``Strategies for pre-training graph neural networks,'' \emph{International
  Conference on Learning Representations}, 2020.

\bibitem{yehudai2021local}
G.~Yehudai, E.~Fetaya, E.~Meirom, G.~Chechik, and H.~Maron, ``From local
  structures to size generalization in graph neural networks,'' in
  \emph{International Conference on Machine Learning}.\hskip 1em plus 0.5em
  minus 0.4em\relax PMLR, 2021, pp. 11\,975--11\,986.

\bibitem{liu2022confidence}
H.~Liu, B.~Hu, X.~Wang, C.~Shi, Z.~Zhang, and J.~Zhou, ``Confidence may cheat:
  Self-training on graph neural networks under distribution shift,'' \emph{The
  Web Conference}, 2022.

\bibitem{li2022let}
S.~Li, X.~Wang, A.~Zhang, Y.~Wu, X.~He, and T.-S. Chua, ``Let invariant
  rationale discovery inspire graph contrastive learning,'' in
  \emph{International Conference on Machine Learning}.\hskip 1em plus 0.5em
  minus 0.4em\relax PMLR, 2022, pp. 13\,052--13\,065.

\bibitem{chen2022graphtta}
G.~Chen, J.~Zhang, X.~Xiao, and Y.~Li, ``Graphtta: Test time adaptation on
  graph neural networks,'' \emph{arXiv preprint arXiv:2208.09126}, 2022.

\bibitem{wang2022test}
Y.~Wang, C.~Li, W.~Jin, R.~Li, J.~Zhao, J.~Tang, and X.~Xie, ``Test-time
  training for graph neural networks,'' \emph{arXiv preprint arXiv:2210.08813},
  2022.

\bibitem{hastings1970monte}
W.~K. Hastings, ``Monte carlo sampling methods using markov chains and their
  applications,'' 1970.

\bibitem{mcpherson2001birds}
M.~McPherson, L.~Smith-Lovin, and J.~M. Cook, ``Birds of a feather: Homophily
  in social networks,'' \emph{Annual review of sociology}, pp. 415--444, 2001.

\bibitem{shafahi2019adversarial}
A.~Shafahi, M.~Najibi, M.~A. Ghiasi, Z.~Xu, J.~Dickerson, C.~Studer, L.~S.
  Davis, G.~Taylor, and T.~Goldstein, ``Adversarial training for free!''
  \emph{Neural Information Processing Systems}, vol.~32, 2019.

\bibitem{ding2021closer}
M.~Ding, K.~Kong, J.~Chen, J.~Kirchenbauer, M.~Goldblum, D.~Wipf, F.~Huang, and
  T.~Goldstein, ``A closer look at distribution shifts and out-of-distribution
  generalization on graphs,'' \emph{NeurIPS Workshop}, 2021.

\bibitem{verma2019manifold}
V.~Verma, A.~Lamb, C.~Beckham, A.~Najafi, I.~Mitliagkas, D.~Lopez-Paz, and
  Y.~Bengio, ``Manifold mixup: Better representations by interpolating hidden
  states,'' in \emph{International Conference on Machine Learning}.\hskip 1em
  plus 0.5em minus 0.4em\relax PMLR, 2019, pp. 6438--6447.

\bibitem{zhang2020does}
L.~Zhang, Z.~Deng, K.~Kawaguchi, A.~Ghorbani, and J.~Zou, ``How does mixup help
  with robustness and generalization?'' in \emph{International Conference on
  Learning Representations}, 2021.

\bibitem{guo2020nonlinear}
H.~Guo, ``Nonlinear mixup: Out-of-manifold data augmentation for text
  classification,'' in \emph{Association for the Advancement of Artificial
  Intelligence}, vol.~34, no.~04, 2020, pp. 4044--4051.

\bibitem{verma2021graphmix}
V.~Verma, M.~Qu, K.~Kawaguchi, A.~Lamb, Y.~Bengio, J.~Kannala, and J.~Tang,
  ``Graphmix: Improved training of gnns for semi-supervised learning,'' in
  \emph{Association for the Advancement of Artificial Intelligence}, vol.~35,
  no.~11, 2021, pp. 10\,024--10\,032.

\bibitem{wang2021mixup}
Y.~Wang, W.~Wang, Y.~Liang, Y.~Cai, and B.~Hooi, ``Mixup for node and graph
  classification,'' in \emph{Proceedings of the Web Conference 2021}, 2021, pp.
  3663--3674.

\bibitem{wu2021graphmixup}
L.~Wu, H.~Lin, Z.~Gao, C.~Tan, S.~Li \emph{et~al.}, ``Graphmixup: Improving
  class-imbalanced node classification on graphs by self-supervised context
  prediction,'' \emph{arXiv preprint arXiv:2106.11133}, 2021.

\bibitem{guo2021intrusion}
H.~Guo and Y.~Mao, ``Intrusion-free graph mixup,'' \emph{arXiv preprint
  arXiv:2110.09344}, 2021.

\bibitem{wang2020nodeaug}
Y.~Wang, W.~Wang, Y.~Liang, Y.~Cai, J.~Liu, and B.~Hooi, ``Nodeaug:
  Semi-supervised node classification with data augmentation,'' in
  \emph{Proceedings of the 26th ACM SIGKDD International Conference on
  Knowledge Discovery \& Data Mining}, 2020, pp. 207--217.

\bibitem{han2022g}
X.~Han, Z.~Jiang, N.~Liu, and X.~Hu, ``G-mixup: Graph data augmentation for
  graph classification,'' \emph{International Conference on Machine Learning},
  2022.

\bibitem{wang2022disentangled}
X.~Wang, H.~Chen, S.~Tang, Z.~Wu, and W.~Zhu, ``Disentangled representation
  learning,'' \emph{arXiv preprint arXiv:2211.11695}, 2022.

\bibitem{montero2020role}
M.~L. Montero, C.~J. Ludwig, R.~P. Costa, G.~Malhotra, and J.~Bowers, ``The
  role of disentanglement in generalisation,'' in \emph{International
  Conference on Learning Representations}, 2020.

\bibitem{dittadi2020transfer}
A.~Dittadi, F.~Tr{\"a}uble, F.~Locatello, M.~W{\"u}thrich, V.~Agrawal,
  O.~Winther, S.~Bauer, and B.~Sch{\"o}lkopf, ``On the transfer of disentangled
  representations in realistic settings,'' \emph{arXiv preprint
  arXiv:2010.14407}, 2020.

\bibitem{gretton2007kernel}
A.~Gretton, K.~Fukumizu, C.~Teo, L.~Song, B.~Sch{\"o}lkopf, and A.~Smola, ``A
  kernel statistical test of independence,'' \emph{Advances in neural
  information processing systems}, vol.~20, 2007.

\bibitem{jaiswal2020survey}
A.~Jaiswal, A.~R. Babu, M.~Z. Zadeh, D.~Banerjee, and F.~Makedon, ``A survey on
  contrastive self-supervised learning,'' \emph{Technologies}, vol.~9, no.~1,
  p.~2, 2020.

\bibitem{le2020contrastive}
P.~H. Le-Khac, G.~Healy, and A.~F. Smeaton, ``Contrastive representation
  learning: A framework and review,'' \emph{IEEE Access}, vol.~8, pp.
  193\,907--193\,934, 2020.

\bibitem{scholkopf2021toward}
B.~Sch{\"o}lkopf, F.~Locatello, S.~Bauer, N.~R. Ke, N.~Kalchbrenner, A.~Goyal,
  and Y.~Bengio, ``Toward causal representation learning,'' \emph{Proceedings
  of the IEEE}, vol. 109, no.~5, pp. 612--634, 2021.

\bibitem{kuang2018stable}
K.~Kuang, P.~Cui, S.~Athey, R.~Xiong, and B.~Li, ``Stable prediction across
  unknown environments,'' in \emph{Proceedings of the 24th ACM SIGKDD
  international conference on knowledge discovery \& data mining}, 2018, pp.
  1617--1626.

\bibitem{rahimi2007random}
A.~Rahimi and B.~Recht, ``Random features for large-scale kernel machines,''
  \emph{Advances in neural information processing systems}, vol.~20, 2007.

\bibitem{xu2018powerful}
K.~Xu, W.~Hu, J.~Leskovec, and S.~Jegelka, ``How powerful are graph neural
  networks?'' in \emph{International Conference on Learning Representations},
  2019.

\bibitem{ying2018hierarchical}
Z.~Ying, J.~You, C.~Morris, X.~Ren, W.~Hamilton, and J.~Leskovec,
  ``Hierarchical graph representation learning with differentiable pooling,''
  \emph{Advances in neural information processing systems}, vol.~31, 2018.

\bibitem{glymour2016causal}
M.~Glymour, J.~Pearl, and N.~P. Jewell, \emph{Causal inference in statistics: A
  primer}.\hskip 1em plus 0.5em minus 0.4em\relax John Wiley \& Sons, 2016.

\bibitem{balke2011probabilistic}
A.~Balke and J.~Pearl, ``Probabilistic evaluation of counterfactual queries,''
  \emph{Association for the Advancement of Artificial Intelligence}, 1994.

\bibitem{pearl2018book}
J.~Pearl and D.~Mackenzie, \emph{The book of why: the new science of cause and
  effect}.\hskip 1em plus 0.5em minus 0.4em\relax Basic books, 2018.

\bibitem{granger1969investigating}
C.~W. Granger, ``Investigating causal relations by econometric models and
  cross-spectral methods,'' \emph{Econometrica: journal of the Econometric
  Society}, pp. 424--438, 1969.

\bibitem{arjovsky2019invariant}
M.~Arjovsky, L.~Bottou, I.~Gulrajani, and D.~Lopez-Paz, ``Invariant risk
  minimization,'' \emph{arXiv preprint arXiv:1907.02893}, 2019.

\bibitem{chang2020invariant}
S.~Chang, Y.~Zhang, M.~Yu, and T.~Jaakkola, ``Invariant rationalization,'' in
  \emph{International Conference on Machine Learning}.\hskip 1em plus 0.5em
  minus 0.4em\relax PMLR, 2020, pp. 1448--1458.

\bibitem{ahuja2021invariance}
K.~Ahuja, E.~Caballero, D.~Zhang, Y.~Bengio, I.~Mitliagkas, and I.~Rish,
  ``Invariance principle meets information bottleneck for out-of-distribution
  generalization,'' \emph{Neural Information Processing Systems (NeurIPS)},
  2021.

\bibitem{tishby2015deep}
N.~Tishby and N.~Zaslavsky, ``Deep learning and the information bottleneck
  principle,'' in \emph{2015 ieee information theory workshop (itw)}.\hskip 1em
  plus 0.5em minus 0.4em\relax IEEE, 2015, pp. 1--5.

\bibitem{kipf2016semi}
T.~N. Kipf and M.~Welling, ``Semi-supervised classification with graph
  convolutional networks,'' in \emph{International Conference on Learning
  Representations}, 2017.

\bibitem{long2015learning}
M.~Long, Y.~Cao, J.~Wang, and M.~Jordan, ``Learning transferable features with
  deep adaptation networks,'' in \emph{International conference on machine
  learning}.\hskip 1em plus 0.5em minus 0.4em\relax PMLR, 2015, pp. 97--105.

\bibitem{zellinger2019robust}
W.~Zellinger, B.~A. Moser, T.~Grubinger, E.~Lughofer, T.~Natschl{\"a}ger, and
  S.~Saminger-Platz, ``Robust unsupervised domain adaptation for neural
  networks via moment alignment,'' \emph{Information Sciences}, vol. 483, pp.
  174--191, 2019.

\bibitem{wu2019simplifying}
F.~Wu, A.~Souza, T.~Zhang, C.~Fifty, T.~Yu, and K.~Weinberger, ``Simplifying
  graph convolutional networks,'' in \emph{International conference on machine
  learning}.\hskip 1em plus 0.5em minus 0.4em\relax PMLR, 2019, pp. 6861--6871.

\bibitem{gretton2009covariate}
A.~Gretton, A.~Smola, J.~Huang, M.~Schmittfull, K.~Borgwardt, and
  B.~Sch{\"o}lkopf, ``Covariate shift by kernel mean matching,'' \emph{Dataset
  shift in machine learning}, vol.~3, no.~4, p.~5, 2009.

\bibitem{ganin2016domain}
Y.~Ganin, E.~Ustinova, H.~Ajakan, P.~Germain, H.~Larochelle, F.~Laviolette,
  M.~Marchand, and V.~Lempitsky, ``Domain-adversarial training of neural
  networks,'' \emph{The journal of machine learning research}, vol.~17, no.~1,
  pp. 2096--2030, 2016.

\bibitem{rahimian2019distributionally}
H.~Rahimian and S.~Mehrotra, ``Distributionally robust optimization: A
  review,'' \emph{arXiv preprint arXiv:1908.05659}, 2019.

\bibitem{dou2019domain}
Q.~Dou, D.~Coelho~de Castro, K.~Kamnitsas, and B.~Glocker, ``Domain
  generalization via model-agnostic learning of semantic features,''
  \emph{Neural Information Processing Systems}, vol.~32, 2019.

\bibitem{mahajan2021domain}
D.~Mahajan, S.~Tople, and A.~Sharma, ``Domain generalization using causal
  matching,'' in \emph{International Conference on Machine Learning}.\hskip 1em
  plus 0.5em minus 0.4em\relax PMLR, 2021, pp. 7313--7324.

\bibitem{zhang2022correct}
M.~Zhang, N.~S. Sohoni, H.~R. Zhang, C.~Finn, and C.~R{\'e},
  ``Correct-n-contrast: A contrastive approach for improving robustness to
  spurious correlations,'' \emph{arXiv preprint arXiv:2203.01517}, 2022.

\bibitem{you2020does}
Y.~You, T.~Chen, Z.~Wang, and Y.~Shen, ``When does self-supervision help graph
  convolutional networks?'' in \emph{international conference on machine
  learning}.\hskip 1em plus 0.5em minus 0.4em\relax PMLR, 2020, pp.
  10\,871--10\,880.

\bibitem{xu2021neural}
K.~Xu, M.~Zhang, J.~Li, S.~S. Du, K.-i. Kawarabayashi, and S.~Jegelka, ``How
  neural networks extrapolate: From feedforward to graph neural networks,''
  \emph{International Conference on Learning Representations}, 2021.

\bibitem{knyazev2019understanding}
B.~Knyazev, G.~W. Taylor, and M.~Amer, ``Understanding attention and
  generalization in graph neural networks,'' \emph{Advances in neural
  information processing systems}, vol.~32, 2019.

\bibitem{gui2022good}
S.~Gui, X.~Li, L.~Wang, and S.~Ji, ``{GOOD}: A graph out-of-distribution
  benchmark,'' in \emph{Neural Information Processing Systems Datasets and
  Benchmarks Track}, 2022.

\bibitem{yuan2022explainability}
H.~Yuan, H.~Yu, S.~Gui, and S.~Ji, ``Explainability in graph neural networks: A
  taxonomic survey,'' \emph{IEEE Transactions on Pattern Analysis and Machine
  Intelligence}, 2022.

\bibitem{ji2022drugood}
Y.~Ji, L.~Zhang, J.~Wu, B.~Wu, L.-K. Huang, T.~Xu, Y.~Rong, L.~Li, J.~Ren,
  D.~Xue \emph{et~al.}, ``Drugood: Out-of-distribution (ood) dataset curator
  and benchmark for ai-aided drug discovery--a focus on affinity prediction
  problems with noise annotations,'' \emph{arXiv preprint arXiv:2201.09637},
  2022.

\bibitem{rozemberczki2021multi}
B.~Rozemberczki, C.~Allen, and R.~Sarkar, ``Multi-scale attributed node
  embedding,'' \emph{Journal of Complex Networks}, vol.~9, no.~2, p. cnab014,
  2021.

\bibitem{pareja2020evolvegcn}
A.~Pareja, G.~Domeniconi, J.~Chen, T.~Ma, T.~Suzumura, H.~Kanezashi, T.~Kaler,
  T.~Schardl, and C.~Leiserson, ``Evolvegcn: Evolving graph convolutional
  networks for dynamic graphs,'' in \emph{Association for the Advancement of
  Artificial Intelligence}, vol.~34, no.~04, 2020, pp. 5363--5370.

\bibitem{scarselli2018vapnik}
F.~Scarselli, A.~C. Tsoi, and M.~Hagenbuchner, ``The vapnik--chervonenkis
  dimension of graph and recursive neural networks,'' \emph{Neural Networks},
  vol. 108, pp. 248--259, 2018.

\bibitem{vapnik2015uniform}
V.~N. Vapnik and A.~Y. Chervonenkis, ``On the uniform convergence of relative
  frequencies of events to their probabilities,'' in \emph{Measures of
  complexity}.\hskip 1em plus 0.5em minus 0.4em\relax Springer, 2015, pp.
  11--30.

\bibitem{verma2019stability}
S.~Verma and Z.-L. Zhang, ``Stability and generalization of graph convolutional
  neural networks,'' in \emph{Proceedings of the 25th ACM SIGKDD International
  Conference on Knowledge Discovery \& Data Mining}, 2019, pp. 1539--1548.

\bibitem{bousquet2002stability}
O.~Bousquet and A.~Elisseeff, ``Stability and generalization,'' \emph{The
  Journal of Machine Learning Research}, vol.~2, pp. 499--526, 2002.

\bibitem{garg2020generalization}
V.~Garg, S.~Jegelka, and T.~Jaakkola, ``Generalization and representational
  limits of graph neural networks,'' in \emph{International Conference on
  Machine Learning}.\hskip 1em plus 0.5em minus 0.4em\relax PMLR, 2020, pp.
  3419--3430.

\bibitem{lv2021generalization}
S.~Lv, ``Generalization bounds for graph convolutional neural networks via
  rademacher complexity,'' \emph{arXiv preprint arXiv:2102.10234}, 2021.

\bibitem{liao2020pac}
R.~Liao, R.~Urtasun, and R.~Zemel, ``A pac-bayesian approach to generalization
  bounds for graph neural networks,'' \emph{International Conference on
  Learning Representations}, 2021.

\bibitem{ma2021subgroup}
J.~Ma, J.~Deng, and Q.~Mei, ``Subgroup generalization and fairness of graph
  neural networks,'' \emph{Advances in Neural Information Processing Systems},
  vol.~34, 2021.

\bibitem{du2019graph}
S.~S. Du, K.~Hou, R.~R. Salakhutdinov, B.~Poczos, R.~Wang, and K.~Xu, ``Graph
  neural tangent kernel: Fusing graph neural networks with graph kernels,''
  \emph{Advancesf in neural information processing systems}, vol.~32, 2019.

\bibitem{zhang2020fast}
S.~Zhang, M.~Wang, S.~Liu, P.-Y. Chen, and J.~Xiong, ``Fast learning of graph
  neural networks with guaranteed generalizability: one-hidden-layer case,'' in
  \emph{International Conference on Machine Learning}.\hskip 1em plus 0.5em
  minus 0.4em\relax PMLR, 2020, pp. 11\,268--11\,277.

\bibitem{baranwal2021graph}
A.~Baranwal, K.~Fountoulakis, and A.~Jagannath, ``Graph convolution for
  semi-supervised classification: Improved linear separability and
  out-of-distribution generalization,'' \emph{International Conference on
  Machine Learning}, 2021.

\bibitem{maskey2022generalization}
S.~Maskey, R.~Levie, Y.~Lee, and G.~Kutyniok, ``Generalization analysis of
  message passing neural networks on large random graphs,'' in \emph{Advances
  in Neural Information Processing Systems}, 2022.

\bibitem{ying2019gnnexplainer}
R.~Ying, D.~Bourgeois, J.~You, M.~Zitnik, and J.~Leskovec, ``Gnnexplainer:
  Generating explanations for graph neural networks,'' \emph{Advances in neural
  information processing systems}, vol.~32, p. 9240, 2019.

\bibitem{lecun1998gradient}
Y.~LeCun, L.~Bottou, Y.~Bengio, and P.~Haffner, ``Gradient-based learning
  applied to document recognition,'' \emph{Proceedings of the IEEE}, vol.~86,
  no.~11, pp. 2278--2324, 1998.

\bibitem{achanta2012slic}
R.~Achanta, A.~Shaji, K.~Smith, A.~Lucchi, P.~Fua, and S.~S{\"u}sstrunk, ``Slic
  superpixels compared to state-of-the-art superpixel methods,'' \emph{IEEE
  transactions on pattern analysis and machine intelligence}, vol.~34, no.~11,
  pp. 2274--2282, 2012.

\bibitem{devlin2018bert}
J.~Devlin, M.~Chang, K.~Lee, and K.~Toutanova, ``{BERT:} pre-training of deep
  bidirectional transformers for language understanding,'' in \emph{Proceedings
  of the 2019 Conference of the North American Chapter of the Association for
  Computational Linguistics: Human Language Technologies, {NAACL-HLT}}.\hskip
  1em plus 0.5em minus 0.4em\relax Association for Computational Linguistics,
  2019, pp. 4171--4186.

\bibitem{mendez2019chembl}
D.~Mendez, A.~Gaulton, A.~P. Bento, J.~Chambers, M.~De~Veij, E.~F{\'e}lix,
  M.~P. Magari{\~n}os, J.~F. Mosquera, P.~Mutowo, M.~Nowotka \emph{et~al.},
  ``Chembl: towards direct deposition of bioassay data,'' \emph{Nucleic acids
  research}, vol.~47, no.~D1, pp. D930--D940, 2019.

\bibitem{velivckovic2019neural}
P.~Veli{\v{c}}kovi{\'c}, R.~Ying, M.~Padovano, R.~Hadsell, and C.~Blundell,
  ``Neural execution of graph algorithms,'' \emph{International Conference on
  Learning Representations}, 2020.

\bibitem{ko2021learning}
J.~Ko, T.~Kwon, K.~Shin, and J.~Lee, ``Learning to pool in graph neural
  networks for extrapolation,'' \emph{arXiv preprint arXiv:2106.06210}, 2021.

\bibitem{customizedgnn}
Y.~Wang, Y.~Ma, W.~Jin, C.~Li, C.~C. Aggarwal, and J.~Tang, ``Customized graph
  neural networks,'' \emph{arXiv}, 2020.

\bibitem{qin2022graph}
Y.~Qin, X.~Wang, Z.~Zhang, P.~Xie, and W.~Zhu, ``Graph neural architecture
  search under distribution shifts,'' in \emph{International Conference on
  Machine Learning}.\hskip 1em plus 0.5em minus 0.4em\relax PMLR, 2022, pp.
  18\,083--18\,095.

\bibitem{galke2021lifelong}
L.~Galke, B.~Franke, T.~Zielke, and A.~Scherp, ``Lifelong learning of graph
  neural networks for open-world node classification,'' in \emph{2021
  International Joint Conference on Neural Networks (IJCNN)}.\hskip 1em plus
  0.5em minus 0.4em\relax IEEE, 2021, pp. 1--8.

\bibitem{jin2022empowering}
W.~Jin, T.~Zhao, J.~Ding, Y.~Liu, J.~Tang, and N.~Shah, ``Empowering graph
  representation learning with test-time graph transformation,'' \emph{arXiv
  preprint arXiv:2210.03561}, 2022.

\bibitem{han2021reliable}
K.~Han, B.~Lakshminarayanan, and J.~Liu, ``Reliable graph neural networks for
  drug discovery under distributional shift,'' \emph{NeurIPS Workshop}, 2021.

\bibitem{yang2022learning}
N.~Yang, K.~Zeng, Q.~Wu, X.~Jia, and J.~Yan, ``Learning substructure invariance
  for out-of-distribution molecular representations,'' in \emph{Neural
  Information Processing Systems}, 2022.

\bibitem{sinha2020evaluating}
K.~Sinha, S.~Sodhani, J.~Pineau, and W.~L. Hamilton, ``Evaluating logical
  generalization in graph neural networks,'' \emph{International Conference on
  Machine Learning Workshop}, 2020.

\bibitem{li2021does}
R.~Li, Y.~Cao, Q.~Zhu, G.~Bi, F.~Fang, Y.~Liu, and Q.~Li, ``How does knowledge
  graph embedding extrapolate to unseen data: a semantic evidence view,''
  \emph{Association for the Advancement of Artificial Intelligence}, 2022.

\bibitem{li2022critical}
K.~Li, B.~DeCost, K.~Choudhary, M.~Greenwood, and J.~Hattrick-Simpers, ``A
  critical examination of robustness and generalizability of machine learning
  prediction of materials properties,'' \emph{arXiv preprint arXiv:2210.13597},
  2022.

\end{thebibliography}

\begin{IEEEbiography}[{\includegraphics[width=1in,height=1.25in,clip,keepaspectratio]{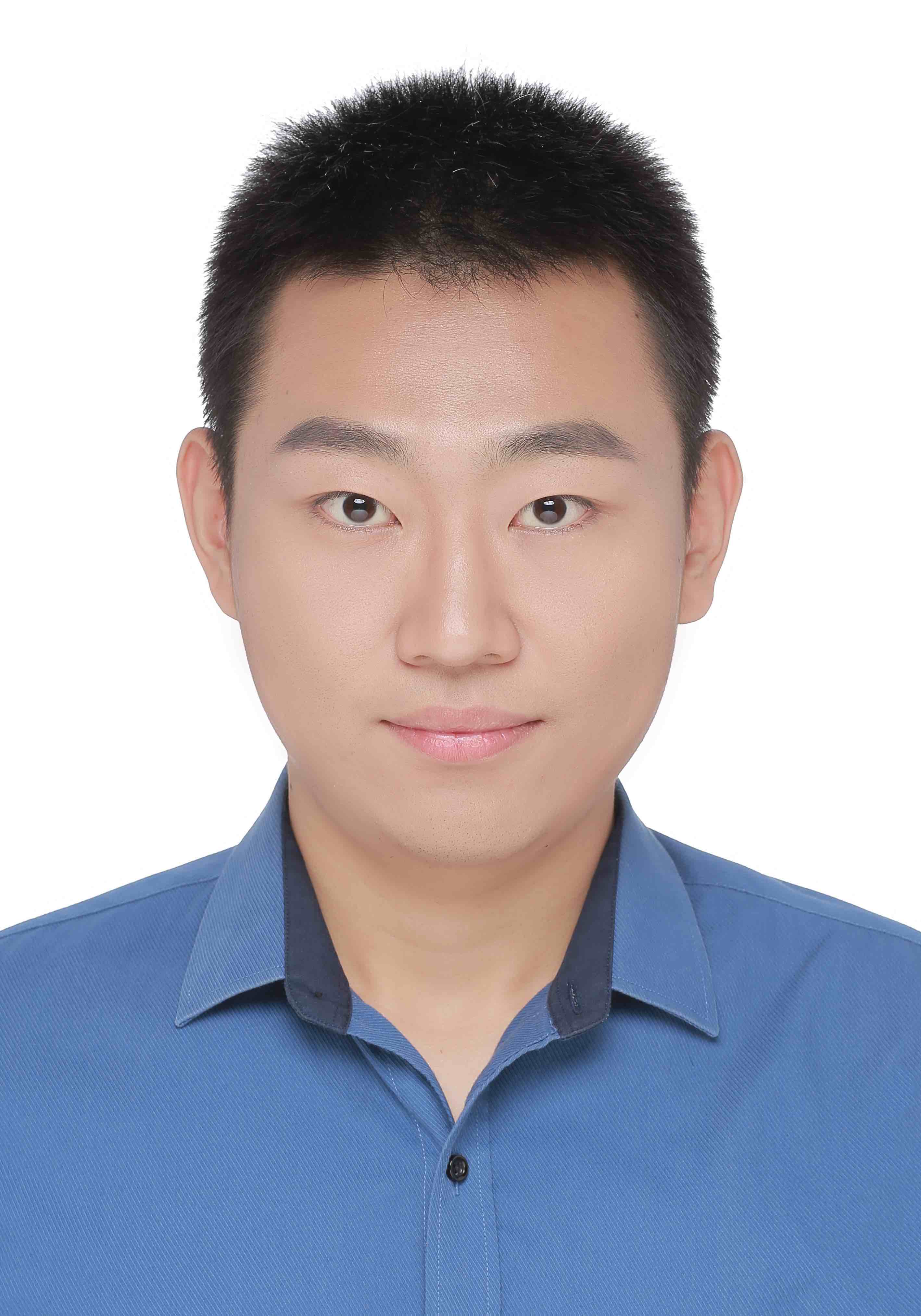}}]{Haoyang Li}
    received his B.E. from the Department of Computer Science and Technology, Tsinghua University in 2018.
    He is a Ph.D. candidate in the Department of Computer Science and Technology of Tsinghua University.
    His research interests are mainly in machine learning on graphs and out-of-distribution generalization.
    He has published several papers in prestigious conferences and journals, e.g., NeurIPS, KDD, ICLR, IEEE TKDE, etc.
\end{IEEEbiography}
    
\begin{IEEEbiography}[{\includegraphics[width=1in,height=1.25in,clip,keepaspectratio]{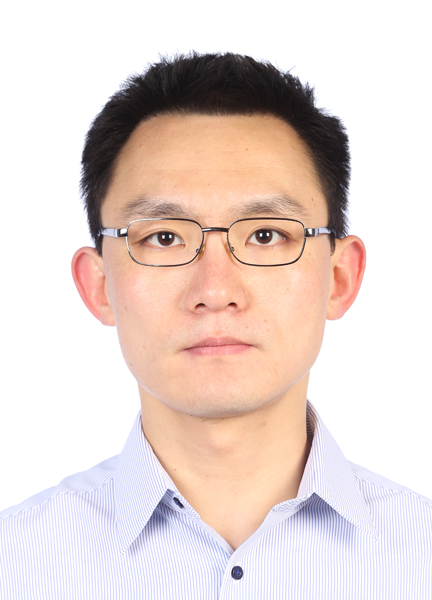}}]{Xin Wang}
is currently an Assistant Professor at the Department of Computer Science and Technology, Tsinghua University. He got both of his Ph.D. and B.E degrees in Computer Science and Technology from Zhejiang University, China. He also holds a Ph.D. degree in Computing Science from Simon Fraser University, Canada. His research interests include multimedia intelligence and recommendation in social media. He has published over 100 high-quality research papers in top conferences and journals including ICML, NeurIPS, IEEE TPAMI, IEEE TKDE, ACM KDD, WWW, ACM SIGIR, ACM Multimedia etc. He is the recipient of 2020 ACM China Rising Star Award and 2022 IEEE TCMC Rising Star Award.
\end{IEEEbiography}

\begin{IEEEbiography}[{\includegraphics[width=1in,height=1.25in,clip,keepaspectratio]{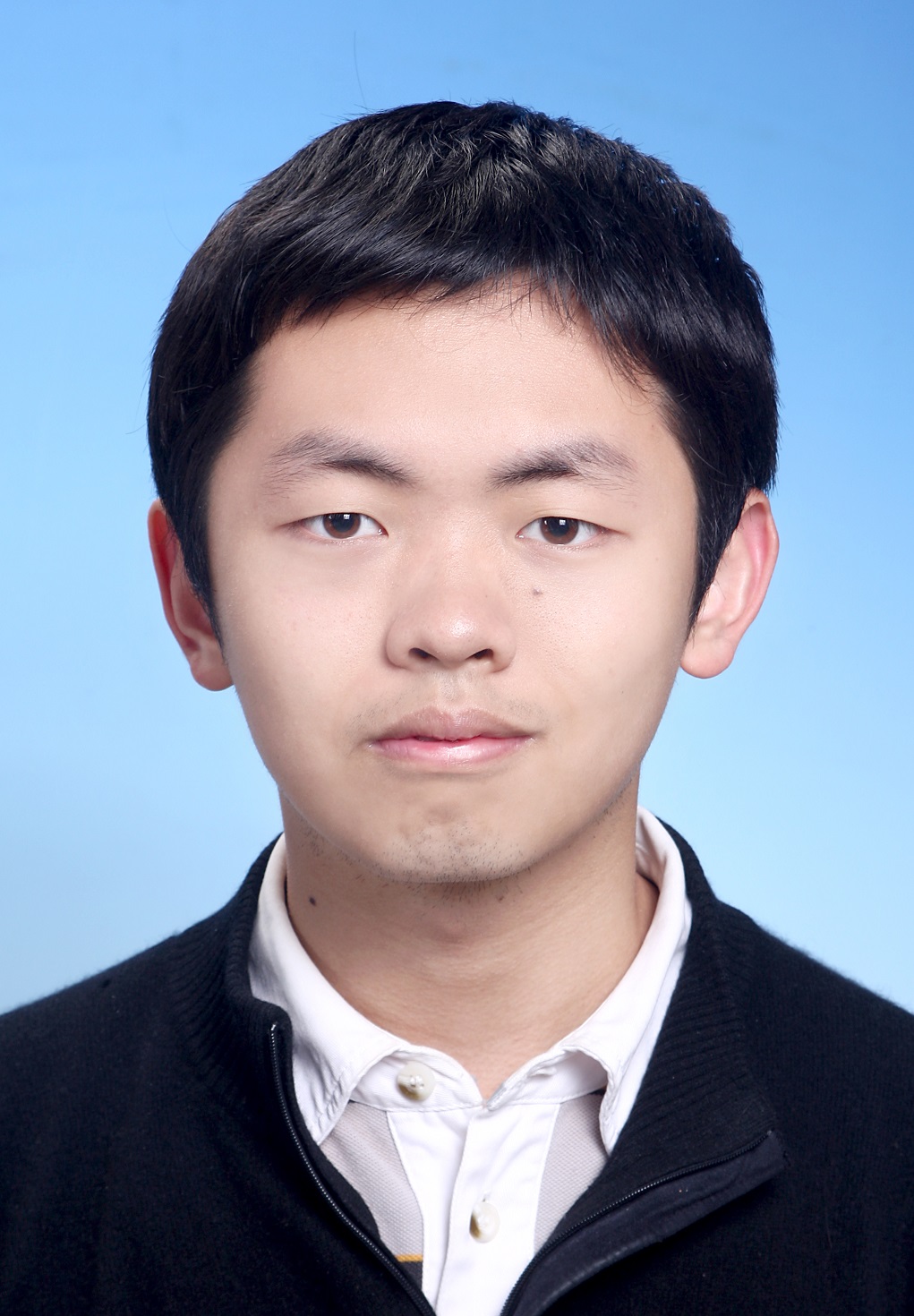}}]{Ziwei Zhang}
received his Ph.D. from the Department of Computer Science and Technology, Tsinghua University, in 2021. He is currently a postdoc researcher in the Department of Computer
Science and Technology at Tsinghua University. His research interests focus on machine learning on graphs, including graph neural network (GNN), network embedding (a.k.a. network representation learning), and automated graph machine learning. He has published over 20 papers in prestigious conferences and journals, including KDD, NeurIPS, ICML, AAAI, IJCAI, and TKDE.
\end{IEEEbiography}

\begin{IEEEbiography}[{\includegraphics[width=1in,height=1.25in,clip,keepaspectratio]{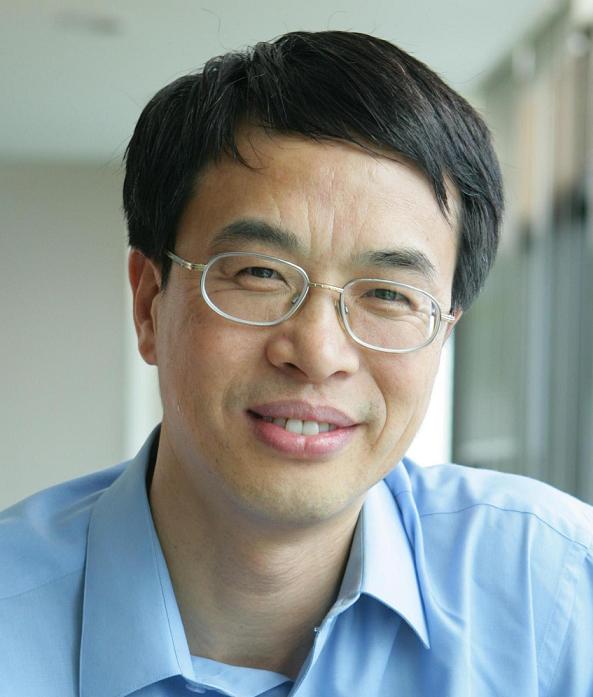}}]{Wenwu Zhu}
    is currently a Professor in the Department of Computer Science and Technology at Tsinghua University, the Vice Dean of National Research Center for Information Science and Technology, and the Vice Director of Tsinghua Center for Big Data. Prior to his current post, he was a Senior Researcher and Research Manager at Microsoft Research Asia. He was the Chief Scientist and Director at Intel Research China from 2004 to 2008. He worked at Bell Labs New Jersey as Member of Technical Staff during 1996-1999. He received his Ph.D. degree from New York University in 1996.
    
    His current research interests are in the area of data-driven multimedia networking and Cross-media big data computing. He has published over 350 referred papers, and is inventor or co-inventor of over 50 patents. He received eight Best Paper Awards, including ACM Multimedia 2012 and IEEE Transactions on Circuits and Systems for Video Technology in 2001 and 2019.  
    
    He served as EiC for IEEE Transactions on Multimedia (2017-2019). He served in the steering committee for IEEE Transactions on Multimedia (2015-2016) and IEEE Transactions on Mobile Computing (2007-2010), respectively. He serves as General Co-Chair for ACM Multimedia 2018 and ACM CIKM 2019, respectively. He is an AAAS Fellow, IEEE Fellow, SPIE Fellow, and a member of The Academy of Europe (Academia Europaea).
\end{IEEEbiography}

\end{document}